\documentclass[10pt,twocolumn,letterpaper]{article}

\usepackage[pagenumbers]{iccv} 

\usepackage{colortbl}

\makeatletter
\renewcommand{\section}{\@startsection{section}{1}{\z@}%
   {-8pt plus -2pt minus -2pt}{5pt}{\large\bf}}
\renewcommand{\subsection}{\@startsection{subsection}{2}{\z@}%
   {-6pt plus -2pt minus -2pt}{3pt}{\elvbf}}
\renewcommand{\subsubsection}{\@startsection{subsubsection}{3}{\z@}%
   {-4pt plus -2pt minus -2pt}{2pt}{\tenbf}}
\makeatother

\AtBeginDocument{%
  \setlength{\abovedisplayskip}{4pt plus 2pt minus 2pt}
  \setlength{\belowdisplayskip}{4pt plus 2pt minus 2pt}
  \setlength{\abovedisplayshortskip}{2pt plus 1pt minus 1pt}
  \setlength{\belowdisplayshortskip}{2pt plus 1pt minus 1pt}
}

\usepackage{graphicx}
\usepackage{amsmath}

\DeclareMathOperator*{\argmin}{arg\,min}
\usepackage{amssymb}
\usepackage{booktabs}

\usepackage{enumitem}

\usepackage{xspace}
\usepackage{amsthm}
\usepackage[numbers]{natbib}
\usepackage[dvipsnames]{xcolor}
\usepackage{adjustbox}
\usepackage{float}
\usepackage[accsupp]{axessibility} 
\usepackage{multirow}
\usepackage{amsfonts}
\usepackage{comment}
\usepackage{makecell}
\usepackage{mathtools}
\usepackage[linesnumbered,ruled,vlined]{algorithm2e}
\usepackage{siunitx} 
\usepackage{tabularx}
\usepackage{pifont} 
\theoremstyle{definition}
\usepackage{subcaption}
\usepackage{pgfplots}

\definecolor{iccvblue}{rgb}{0.21,0.49,0.74}
\usepackage[pagebackref,breaklinks,colorlinks,allcolors=iccvblue]{hyperref}
\usepackage[capitalize]{cleveref}

\crefname{section}{Sec.}{Secs.}
\Crefname{section}{Section}{Sections}
\crefname{table}{Tab.}{Tabs.}
\Crefname{table}{Table}{Tables}

\def\paperID{7035} 
\def\confName{ICCV}
\def\confYear{2025}

\title{Dataset Distillation via the Wasserstein Metric}

\author{
Haoyang Liu\textsuperscript{1}, Yijiang Li\textsuperscript{2}, Tiancheng Xing\textsuperscript{3}, Peiran Wang\textsuperscript{4}, Vibhu Dalal\textsuperscript{5}, Luwei Li\textsuperscript{1} \\
Jingrui He\textsuperscript{1}, Haohan Wang\textsuperscript{1} \\
\\
\textsuperscript{1}University of Illinois at Urbana-Champaign\\
\textsuperscript{2}University of California, San Diego\\
\textsuperscript{3}National University of Singapore\\
\textsuperscript{4}University of California, Los Angeles\\
\textsuperscript{5}Sri Aurobindo International Centre of Education\\
\texttt{\{hl57, luweili2, jingrui, haohanw\}@illinois.edu} \\
}
\date{}

\begin{document}

\newcommand{\sssec}[1]{\vspace*{0.05in}\noindent\textbf{#1}}
\newcolumntype{M}[1]{>{\centering\arraybackslash}p{#1}}

\newcommand{\x}{\mathbf{x}}
\newcommand{\y}{\mathbf{y}}
\newcommand{\X}{\mathbf{X}}
\newcommand{\Y}{\mathbf{Y}}
\newcommand{\Z}{\mathbf{Z}}
\newcommand{\bc}{\mathbf{c}}
\newcommand{\I}{\mathbf{I}}
\newcommand{\B}{\mathbf{B}}
\newcommand{\z}{\mathbf{z}}
\newcommand{\ba}{\mathbf{a}}
\newcommand{\h}{\mathbf{h}}
\newcommand{\bu}{\mathbf{u}}

\def\maketitlesupplementary
   {
   \newpage
   \begingroup 
       \centering
       \Large
       \textbf{\thetitle}\\\vspace{0.5em}
       Supplementary Material\\\vspace{1.0em}
   \endgroup 
   }

\newcommand\blfootnote[1]{%
  \begingroup
  \renewcommand\thefootnote{}\footnote{#1}%
  \addtocounter{footnote}{-1}%
  \endgroup
}

\twocolumn[{%
    \maketitle
    \captionsetup{type=figure} 
\begin{center}
    \begin{adjustbox}{width=\textwidth,center}
        \setlength{\fboxsep}{0pt}
        \begin{subfigure}[b]{0.14\linewidth}
            \centering
            \includegraphics[width=\linewidth]{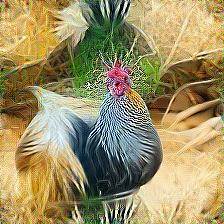}
            \caption*{Cock}
        \end{subfigure}
        \hfill
        \begin{subfigure}[b]{0.14\linewidth}
            \centering
            \includegraphics[width=\linewidth]{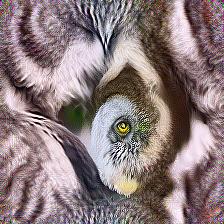}
            \caption*{Grey Owl}
        \end{subfigure}
        \hfill
        \begin{subfigure}[b]{0.14\linewidth}
            \centering
            \includegraphics[width=\linewidth]{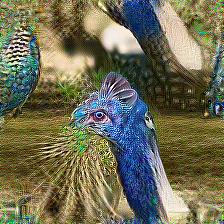}
            \caption*{Peacock}
        \end{subfigure}
        \hfill
        \begin{subfigure}[b]{0.14\linewidth}
            \centering
            \includegraphics[width=\linewidth]{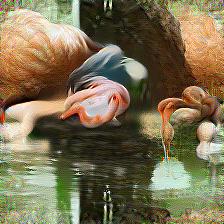}
            \caption*{Flamingo}
        \end{subfigure}
        \hfill
        \begin{subfigure}[b]{0.14\linewidth}
            \centering
            \includegraphics[width=\linewidth]{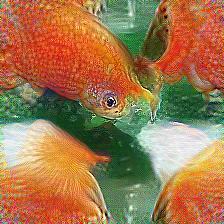}
            \caption*{Gold Fish}
        \end{subfigure}
        \hfill
        \begin{subfigure}[b]{0.14\linewidth}
            \centering
            \includegraphics[width=\linewidth]{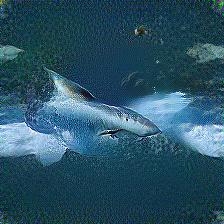}
            \caption*{\scriptsize Shark}
        \end{subfigure}
        \hfill
        \begin{subfigure}[b]{0.14\linewidth}
            \centering
            \includegraphics[width=\linewidth]{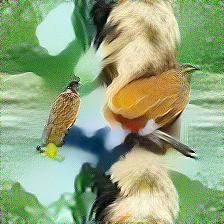}
            \caption*{\scriptsize Bulbul}
        \end{subfigure}
    \end{adjustbox}

    \begin{adjustbox}{width=\linewidth,center}
        \setlength{\fboxsep}{0pt}
        \begin{subfigure}[b]{0.14\linewidth}
            \centering
            \includegraphics[width=\linewidth]{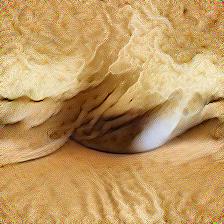}
            \caption*{Dough}
        \end{subfigure}
        \hfill
        \begin{subfigure}[b]{0.14\linewidth}
            \centering
            \includegraphics[width=\linewidth]{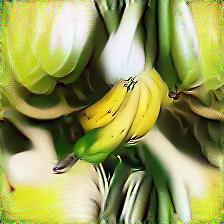}
            \caption*{Banana}
        \end{subfigure}
        \hfill
        \begin{subfigure}[b]{0.14\linewidth}
            \centering
            \includegraphics[width=\linewidth]{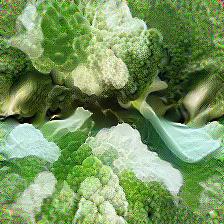}
            \caption*{Broccoli}
        \end{subfigure}
        \hfill
        \begin{subfigure}[b]{0.14\linewidth}
            \centering
            \includegraphics[width=\linewidth]{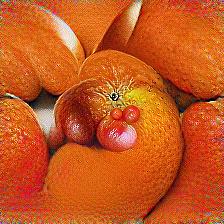}
            \caption*{Orange}
        \end{subfigure}
        \hfill
        \begin{subfigure}[b]{0.14\linewidth}
            \centering
            \includegraphics[width=\linewidth]{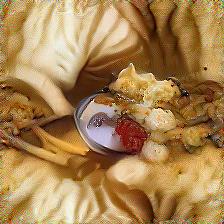}
            \caption*{Potato}
        \end{subfigure}
        \hfill
        \begin{subfigure}[b]{0.14\linewidth}
            \centering
            \includegraphics[width=\linewidth]{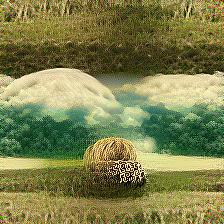}
            \caption*{\scriptsize Hay}
        \end{subfigure}
        \hfill
        \begin{subfigure}[b]{0.14\linewidth}
            \centering
            \includegraphics[width=\linewidth]{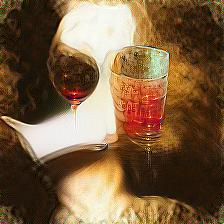}
            \caption*{\scriptsize Red Wine}
        \end{subfigure}
    \end{adjustbox}
    \setcounter{figure}{0}
    \captionof{figure}{Synthetic images distilled from ImageNet-1K using our WMDD method with ResNet-18, capturing essential class features aligned with human perception. We randomly sampled one image for each of the chosen categories from our output in the 10 IPC setting.}
    \label{fig:visualization-example}
\end{center}
}]

\begin{abstract}
Dataset Distillation (DD) aims to generate a compact synthetic dataset that enables models to achieve performance comparable to 
training on the full large dataset, significantly reducing computational costs. Drawing from optimal transport theory, we introduce WMDD 
(Wasserstein Metric-based Dataset Distillation), a straightforward yet powerful method that employs 
the Wasserstein metric to enhance distribution matching.

We compute the Wasserstein barycenter of features from a pretrained classifier to capture essential characteristics of the original data
 distribution. By optimizing synthetic data to align with this barycenter in feature space and leveraging per-class BatchNorm statistics 
 to preserve intra-class variations, WMDD maintains the efficiency of distribution matching approaches while achieving state-of-the-art 
 results across various high-resolution datasets. Our extensive experiments demonstrate WMDD's effectiveness and adaptability, 
 highlighting its potential for advancing machine learning applications at scale. Code is available at \url{https://github.com/Liu-Hy/WMDD} and website at \url{https://liu-hy.github.io/WMDD/}.
\end{abstract}

\section{Introduction}
\label{sec:intro}

Dataset distillation \citep{wang2018dataset, zhao2021DC} aims to create compact synthetic datasets that train models to perform similarly to those trained on full-sized original datasets. This technique promises to address the escalating computational costs associated with growing data volumes, enables efficient model development across various applications \citep{zhang2022dense, goetz2020federated, li2022more, such2020generative, sachdeva2023data, li2023towards}, and helps mitigate bias \citep{xue2024towards, cui2024mitigating}, robustness \cite{xue2025towards} and privacy \cite{dong2022privacy, shiprivacy} concerns in training data.
\begin{figure*}
    \begin{subfigure}{0.37\textwidth}
        \centering
        \includegraphics[width=\textwidth]{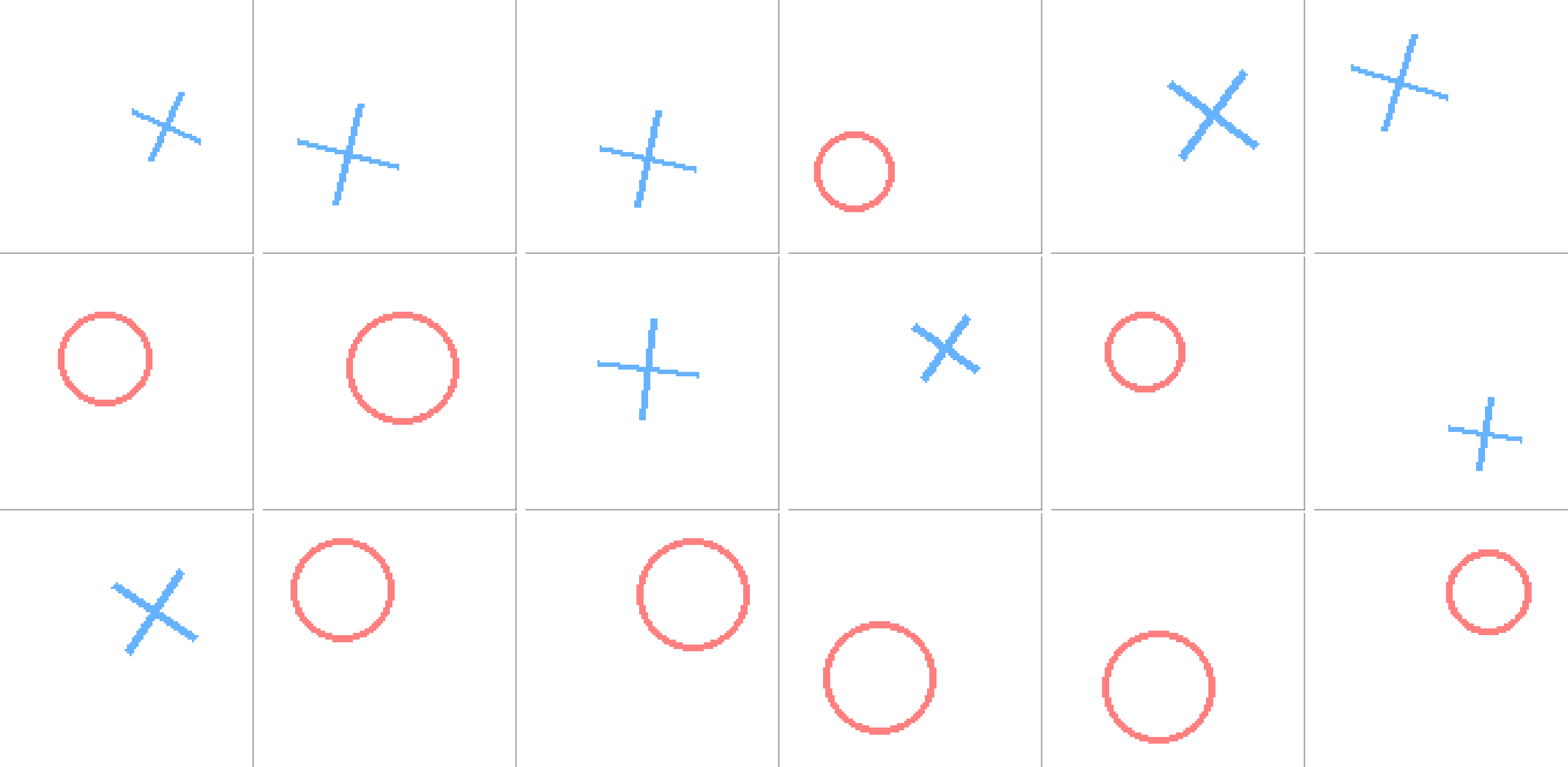}
        \caption{}
        \label{subfig:subfigure1}
    \end{subfigure}
    \hspace{0.03\textwidth}
    \begin{subfigure}{0.16\textwidth}
        \centering
        \fbox{\includegraphics[width=\textwidth]{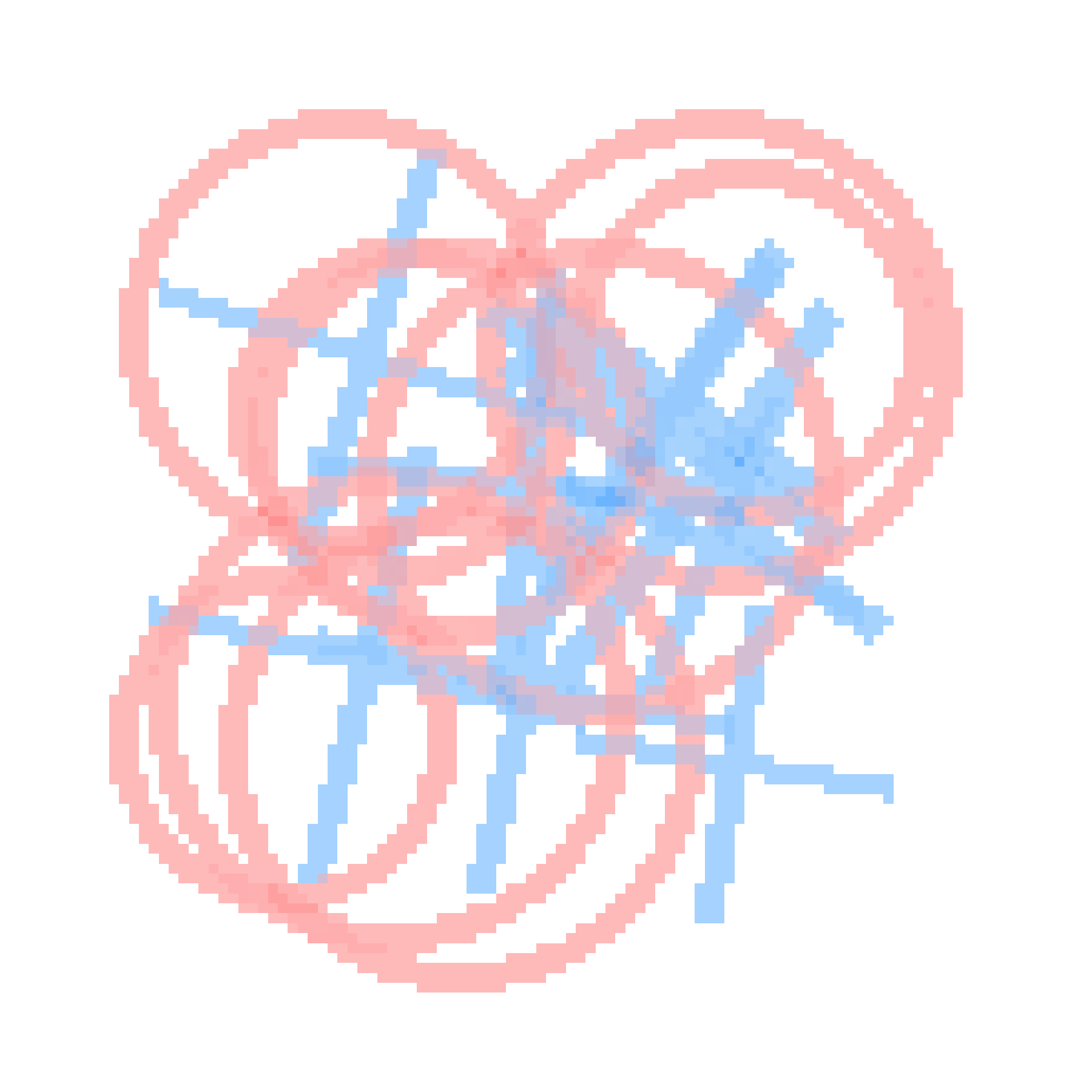}}
        \caption{KL divergence}
        \label{fig:kl}
    \end{subfigure}%
    \hspace{0.03\textwidth}
    \begin{subfigure}{0.16\textwidth}
        \centering
        \fbox{\includegraphics[width=\textwidth]{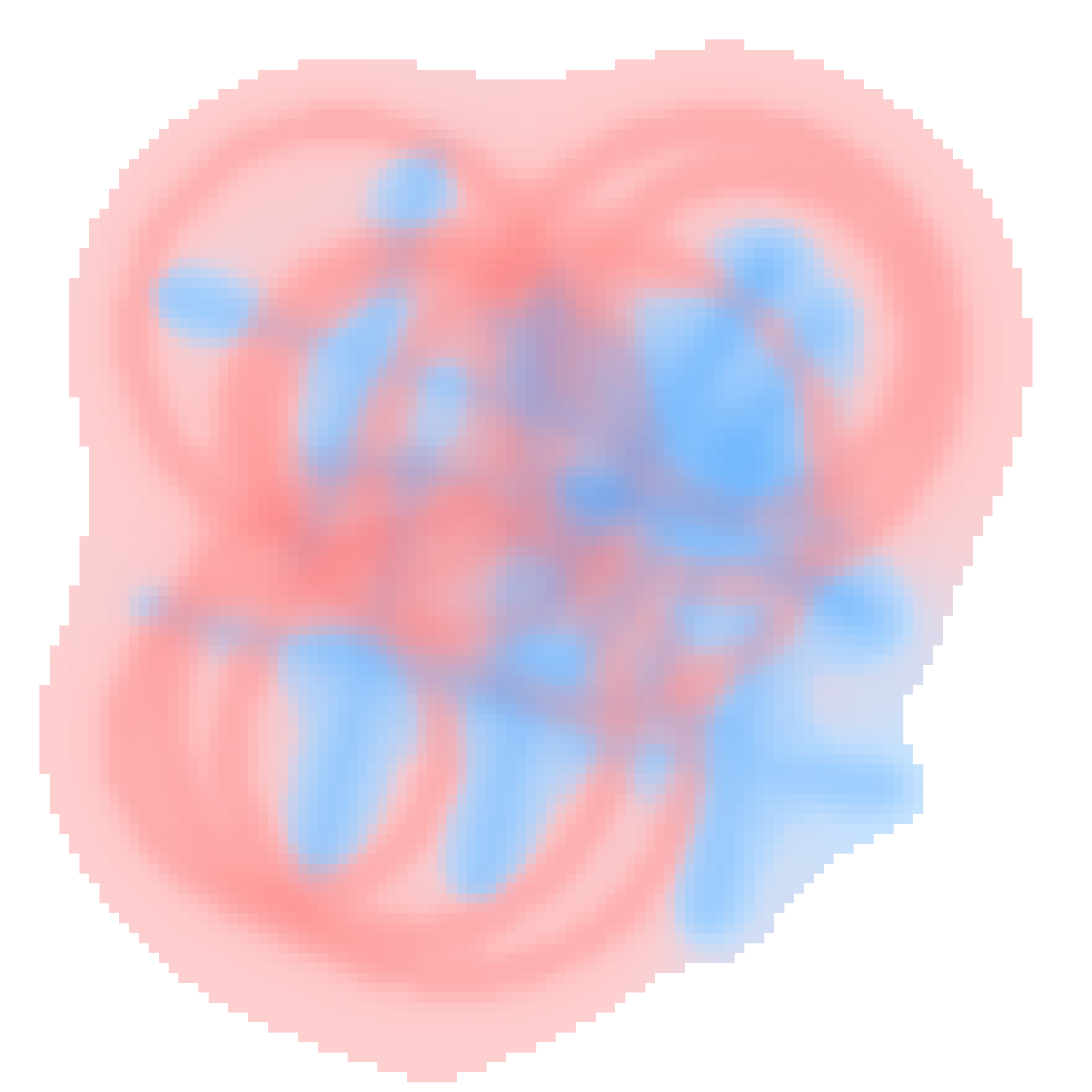}}
        \caption{MMD distance}
        \label{fig:mmd}
    \end{subfigure}%
    \hspace{0.03\textwidth}
    \begin{subfigure}{0.16\textwidth}
        \centering
        \fbox{\includegraphics[width=\textwidth]{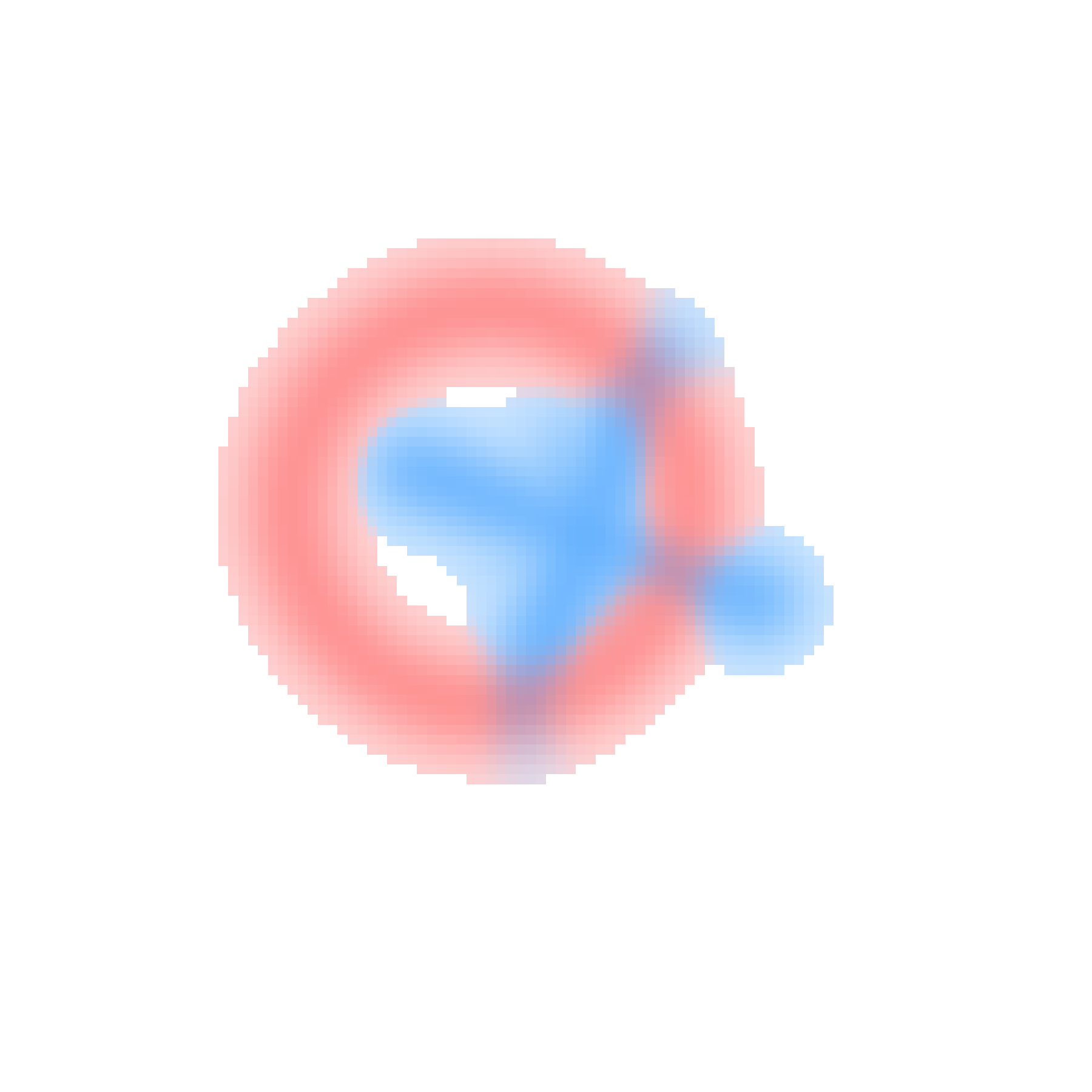}}
        \caption{Wasserstein distance}
        \label{fig:wass}
    \end{subfigure}
    \caption{The capability of Wasserstein barycenter in condensing the core characteristics of distributions: (a) distributions defined on $\mathbb{R}^2$, concentrated on outlines of circles (blue) and crosses (green). Barycenters computed using: (b) KL divergence, (c) Maximum Mean Discrepancy (MMD), which operates in a kernel-induced feature space, and (d) Wasserstein distance, which preserves geometric structure through optimal transport. Color intensity represents probability density, while color hue shows different types of source distributions.}
    \label{fig: distribution_teaser}
\end{figure*}

The central challenge in dataset distillation lies in capturing the distributional characteristics of an entire dataset within a small set of synthetic samples \cite{lei2022comprehensive, sachdeva2023data}. Existing methods often struggle to balance computational efficiency with distillation quality. Some researchers formulate dataset distillation as a bi-level optimization problem \citep{wang2020dataset, loo2022efficient, nguyen2021dataset}, which has inspired innovative approaches such as gradient matching \citep{zhao2021DC, zhao2021DSA}, trajectory matching \citep{cazenavette2022distillation}, and curvature matching \citep{shin2023losscurvature}. These methods align the optimization dynamics between models trained on synthetic and original datasets. However, they typically require second-order derivative computation, becoming prohibitively expensive for large datasets like ImageNet-1K \citep{deng2009imagenet}. Alternative approaches directly align synthetic and original data distributions using metrics like Maximum Mean Discrepancy (MMD) \citep{gretton2012kernel, tolstikhin2016minimax}. Despite their computational efficiency, these methods typically underperform compared to optimization-based approaches \cite{sachdeva2023data, lei2022comprehensive}. We conjecture that this performance gap is due to MMD's limitations in quantifying distributional differences in ways that provide meaningful signals for generating effective synthetic images.

In this paper, we introduce the Wasserstein distance as an effective measure of distributional differences for Dataset Distillation. Wasserstein distance is known for comparing distributions by quantifying the minimal movement required to transform one probability distribution into another within a given metric space \citep{villani2008optimal}. Grounded in Optimal Transport theory \citep{kantorovich1960mathematical}, it provides a geometrically meaningful approach to quantifying differences between distributions. The Wasserstein barycenter \citep{agueh2011barycenters} represents the centroid of multiple distributions while preserving their essential characteristics. \cref{fig: distribution_teaser} illustrates this advantage by simulating distributions spread on circles and crosses on a 2D plane (\cref{subfig:subfigure1}), and their barycenters computed with different distribution metrics. While KL divergence (\cref{fig:kl}) and MMD (\cref{fig:mmd}) barycenters produce a rigid mix-up of input distributions, the Wasserstein barycenter (\cref{fig:wass}) creates a natural interpolation that preserves the structural characteristics of the original distributions.

Motivated by these advantages, we develop a straightforward yet effective DD method using Wasserstein distance for distribution matching. Unlike prior work using MMD \citep{gretton2012kernel, tolstikhin2016minimax}, the Wasserstein barycenter \citep{agueh2011barycenters} avoids reliance on heuristically designed kernels and naturally accounts for distribution geometry and structure. This allows us to statistically summarize real datasets within a fixed number of representative and diverse synthetic images that enable classification models to achieve higher performance.

Furthermore, to address challenges in optimizing high-dimensional data for DD, we present WMDD (\textbf{W}asserstein \textbf{M}etric-based \textbf{D}ataset \textbf{D}istillation), 
an algorithm that balances performance and computational feasibility on large datasets.
We embed synthetic data into the feature space of a pre-trained image classifier following \citep{yin2023squeeze, zhao2023dataset, zhao2023improved}, and use the Wasserstein barycenter as a compact summary of intra-class data distribution. To leverage prior knowledge in pretrained models, we propose a regularization method using Per-Class BatchNorm statistics (PCBN) for more precise distribution matching, inspired by previous work addressing data heterogeneity \cite{gong2022sandwich} and long-tail problems \cite{cheng2022compound} with variants of batch normalization \cite{ioffe2015batch}. By implementing an efficient algorithm \citep{cuturi2014fast} for Wasserstein barycenter computation, our method maintains the efficiency of distribution matching-based approaches \citep{zhao2023dataset} and can scale to large, high-resolution datasets like ImageNet-1K \citep{deng2009imagenet}. Our experiments demonstrate that WMDD achieves state-of-the-art performance across various benchmarks. Our contributions include:

\begin{itemize}
\item{A novel dataset distillation technique that integrates distribution matching with Wasserstein metrics, bridging dataset distillation with insights from optimal transport theory.}

\item{A balanced solution leveraging the computational feasibility of distribution-matching based methods to ensure scalability to large datasets.} 

\item{Comprehensive experimental results across diverse high-resolution datasets demonstrating significant performance improvements over existing methods, highlighting our approach's practical applicability in the big data era.}
\end{itemize}
        
\section{Related work}
\label{sec:related_work}
\subsection{Data Distillation}

Dataset Distillation (DD) aims to create compact synthetic training sets that enable models to achieve performance comparable to those trained on larger original datasets \citep{wang2018datasetdistillation}. Current DD methods fall into three major categories \citep{yu2023dataset}: \emph{Performance Matching} seeks to minimize loss of the synthetic dataset by aligning the performance of models trained on synthetic and original datasets, methods include DD \citep{wang2018datasetdistillation}, FRePo \citep{zhou2022dataset}, AddMem \citep{deng2022remember}, KIP \citep{nguyen2021dataset}, RFAD \citep{loo2022efficient}; \emph{Parameter Matching} is an approach to train two neural networks on the real and synthetic datasets respectively, with the aim to promote similarity in their parameters, methods include DC \citep{zhao2021DC}, DSA \citep{zhao2021DSA}, MTT \citep{cazenavette2022distillation}, HaBa \citep{liu2022dataset}, FTD \citep{du2023minimizing}, TESLA \citep{cui2023scaling}; \emph{Distribution Matching} aims to obtain synthetic data that closely matches the distribution of real data, methods include DM \citep{zhao2023dataset}, IT-GAN \citep{zhao2022synthesizing}, KFS \citep{lee2022dataset}, CAFE \citep{wang2022cafe}, SRe$^2$L \citep{yin2023squeeze}, IDM \citep{zhao2023improved}, G-VBSM \citep{shao2024generalized}, and SCDD \citep{shao2024generalized}.

\subsection{Distribution Matching} 

Distribution Matching (DM) techniques, initially proposed in \citep{zhao2022dataset}, aim to directly align the probability distributions of the original and synthetic datasets \citep{sachdeva2023data, geng2023survey}. The fundamental premise underlying these methods is that when two datasets exhibit similarity based on a specific distribution divergence metric, they lead to comparably trained models \citep{Lei_2024}. DM typically employs parametric encoders for projecting data onto a lower dimensional latent space and approximates the Maximum Mean Discrepancy for assessing distribution mismatch \citep{yin2023squeeze, wang2022cafe, zhao2023improved, zhang2024m3d, sun2023diversity, zhao2022dataset}. Notably, DM avoids reliance on model parameters and bi-level optimization, diverging from gradient and trajectory matching approaches. This distinction reduces memory requirements. However, the empirical evidence so far suggests that DM may underperform compared to the other approaches \citep{Lei_2024, zhang2024m3d}. 

\section{Preliminaries}
\label{sec:prelim}
We introduce the fundamental concepts of Dataset Distillation and Wasserstein barycenters that form the foundation of our approach.

\subsection{Dataset Distillation}

\paragraph{Notations} 
Let $\mathcal{T}=\{(\mathbf{x}_i,y_i)\}_{i=1}^{n}$ be the real training set that contains $n$ distinct input--label pairs and let $\mu_{\mathcal{T}}$ be its empirical distribution, i.e.\ $\mathbf{x}_i\sim\mu_{\mathcal{T}}$. 
Similarly, let $\mathcal{S}=\{(\tilde{\mathbf{x}}_j,\tilde{y}_j)\}_{j=1}^{m}$ be the synthetic set with \emph{at most} $m$ distinct pairs and empirical distribution $\mu_{\mathcal{S}}$. 
Each data point lies in an ambient space $\Omega = \mathbb{R}^{d}$.  
Denote by $\mathbf{X}\in\mathbb{R}^{n\times d}$ and $\tilde{\mathbf{X}}\in\mathbb{R}^{m\times d}$ the matrices that stack the unique positions in $\mathcal{T}$ and $\mathcal{S}$, respectively.  
The probability mass associated with the synthetic samples is stored in the weight vector $\mathbf{w}\in\Delta^{m-1}$, where $w_j$ is the weight of $\tilde{\mathbf{x}}_j$ and $\Delta^{m-1}$ is the $(m-1)$--simplex.  
Consequently, we can compactly write the synthetic dataset as the tuple $\mathcal{S}=(\tilde{\mathbf{X}},\mathbf{w})$. 
Throughout, $\ell(\mathbf{x},y;\boldsymbol{\theta})$ denotes the loss incurred by a model with parameters $\boldsymbol{\theta}$ on a single sample $(\mathbf{x},y)$.

Dataset Distillation (DD) aims at finding the optimal synthetic set $\mathcal{S}^*$ for a given $\mathcal{T}$ by solving a bi-level optimization problem as below:
\begin{gather}
\label{formulation}
    \mathcal{S}^* = \argmin_{\mathcal{S}} \mathop{\mathbb{E}}_{{(\mathbf{x}, y)} \sim \mu_\mathcal{T}} \ell\left(\mathbf{x}, y; \boldsymbol{\theta}(\mathcal{S})\right) \\
    \textrm{subject to}~~ \boldsymbol{\theta}(\mathcal{S}) = \argmin_{\boldsymbol{\theta}} \sum_{i=1}^m \ell(\tilde{\mathbf{x}}_i, \tilde{y}_i; \boldsymbol{\theta}).
\end{gather}
Directly solving the bi-level optimization problem poses significant challenges. As a viable alternative, a prevalent approach \citep{zhao2023dataset, wang2022cafe, sajedi2022datadam, yin2023squeeze} seeks to align the distribution of the synthetic dataset with that of the real dataset. This strategy is based on the assumption that 
\emph{the optimal synthetic dataset should be the one that is distributionally closest to the real dataset subject to a fixed number of synthetic data points}. 
We label this as \textbf{Assumption A1}.
While recent methods \cite{zhao2023dataset, wang2022cafe, zhao2022synthesizing} grounded on this premise have shown promising empirical results, they often struggle to balance strong performance with scalability to large datasets like ImageNet-1K. 

\subsection{Wasserstein barycenters}
\label{sec:wbc}
Our method computes representative features using Wasserstein barycenters \citep{agueh2011barycenters}, extending the concept of ``averaging'' to distributions while respecting their geometric properties. This approach relies on the Wasserstein distance to quantify distributional differences.

\textbf{Definition~1 (Wasserstein distance).}  
Let $(\Omega, D)$ be a metric space and denote by $P(\Omega)$ the set of Borel probability measures on $\Omega$.  For $\mu,\nu\in P(\Omega)$ the $p$-Wasserstein distance is
\begin{align}
  W_p(\mu,\nu) := \Bigl(\inf_{\pi\in\Pi(\mu,\nu)} \!\int_{\Omega^2} D(x,y)^p\,\mathrm d\pi(x,y)\Bigr)^{1/p},
\end{align}
where $\Pi(\mu,\nu)$ is the set of couplings (joint distributions with the prescribed marginals).  Intuitively, $W_p$ measures the minimum ``work''—mass times distance—required to morph $\mu$ into $\nu$; hence it is also known as the earth–mover distance.  

\textbf{Definition~2 (Wasserstein barycenter).}  
Given $N$ distributions $\{\nu_i\}_{i=1}^{N}\subseteq P(\Omega)$, their $p$-Wasserstein barycenter is any solution of
\begin{align}
\label{eq: def_bary}
  \underset{\mu\in P(\Omega)}{\arg\min}\;f(\mu):=\frac{1}{N}\sum_{i=1}^{N} W_p^p(\mu,\nu_i).
\end{align}
The barycenter can be viewed as the ``center of mass'' of the input distributions: it minimizes the average transportation cost (squared when $p{=}2$) to all $u_i$.

\begin{figure*}
    \centering
    \includegraphics[width=\textwidth]{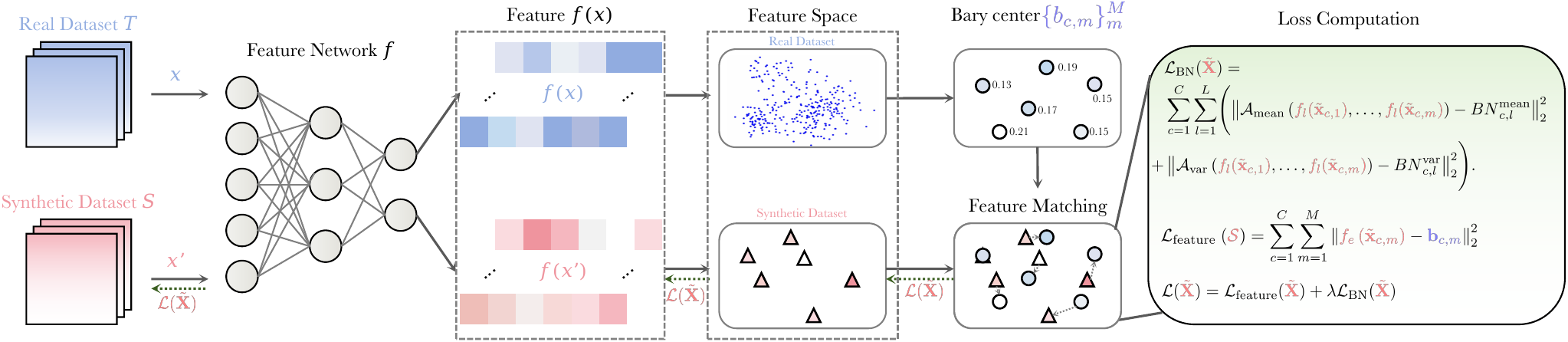}
    \caption{Diagram of our WMDD method. Real dataset $T$ and synthetic dataset $S$ pass through the feature network $f$ to obtain features. The features of the real dataset are used to compute the Wasserstein Barycenter. The synthetic dataset is optimized via feature matching and loss computation (combining feature loss and BN regularization) to align with the Barycenter, generating high-quality synthetic data for efficient model training.}
    \label{fig:wmdd-scheme}
\end{figure*}

\section{Method}
\label{sec:approach}
The Wasserstein distance offers an intuitive and geometrically meaningful way to quantify differences between distributions, as demonstrated by its superior performance in preserving structural characteristics (Fig. \ref{fig: distribution_teaser}). We leverage these strengths to bridge the performance gap in dataset distillation and potentially surpass current state-of-the-art techniques. This section establishes the connection between Wasserstein barycenters and dataset distillation, presents the efficient computation approach, and introduces our complete method design.

\subsection{Wasserstein barycenter in dataset distillation}
We begin by representing both real and synthetic datasets as empirical distributions. For the real dataset $\mathcal{T}$, assuming no prior knowledge and no repetitive samples, we adopt a discrete uniform distribution over the observed data points, 
$\mu_{\mathcal{T}} = \frac{1}{n} \sum_{i=1}^n \delta_{\mathbf{x}_i}$, 
where $\delta_{\mathbf{x}_i}$ represents the Dirac delta function centered at position $\mathbf{x}_i$. This function is zero everywhere except at $\mathbf{x}_i$ and integrates to one.

For the synthetic dataset $\mathcal{S}$, we define its empirical distribution as:
$\mu_{\mathcal{S}} = \sum_{j=1}^m w_j \delta_{\tilde{\mathbf{x}}_j}$,
where the weights satisfy $w_j\ge 0$ and $\sum_{j=1}^m w_j = 1$. Learning these probabilities provides additional flexibility in approximating the real distribution.

Following Assumption A1 and our choice of the Wasserstein metric, the optimal synthetic dataset $\mathcal{S}^*$ should generate an empirical distribution that minimizes the Wasserstein distance to the real data distribution:
\begin{align}
\mu_{\mathcal{S}^*} = \mu_{\mathcal{S}}^* = \argmin_{\mu_{\mathcal{S}} \in P_m} W_p^p(\mu_{\mathcal{S}}, \mu_{\mathcal{T}}),
\label{eq:wb_dd}
\end{align}
where $\mu_{\mathcal{S}^*}$ is the empirical distribution of the optimal dataset, $\mu_{\mathcal{S}}^*$ is the optimal empirical distribution, and $P_m\subset P(\Omega)$ denotes the set of distributions supported on at most $m$ atoms in $\mathbb{R}^d$. This is a special case of \eqref{eq: def_bary} with $N=1$.
Since the synthetic set $\mathcal{S}$ is fully specified by positions $\tilde{\mathbf{X}}$ and weights $\mathbf{w}$, 
we can find the optimal set $\mathcal{S}^*$ by minimizing the below function:
\begin{align}
    f(\tilde{\mathbf{X}}, \mathbf{w}) := W_p^p(\mu_{\mathcal{S}}, \mu_{\mathcal{T}}).
\end{align}

\subsection{Computing the Wasserstein barycenter}
\label{sec: bary_compute}
To efficiently optimize $f(\tilde{\mathbf{X}}, \mathbf{w})$, we adapt the barycenter computation method from \citep{cuturi2014fast}, employing an alternating optimization approach that iterates between optimizing weights and positions. This approach leverages the convex structure of the optimal transport problem to ensure computational efficiency.
\paragraph{Weight optimization with fixed positions}
With fixed synthetic data positions $\tilde{\mathbf{X}}$, we first construct a cost matrix $\mathbf{C} \in \mathbb{R}^{n \times m}$ where each \(c_{ij} = \|\tilde{\mathbf{x}}_j - \mathbf{x}_i\|^2\) represents the squared Euclidean distance between points in the two distributions. The Wasserstein distance calculation transforms into finding the optimal transport plan $\mathbf{T} \in \mathbb{R}^{n \times m}$, where each $t_{ij}$ represents the mass moved from position $i$ to position $j$:
\begin{gather}
\min_{\mathbf{T}} \langle \mathbf{C}, \mathbf{T} \rangle_F \quad  \label{eq: cost} \text{subject to} \quad \sum_{j=1}^{m} t_{ij} = \frac{1}{n}, \; \forall i, \quad \\
\sum_{i=1}^{n} t_{ij} = w_j,  \; \forall j, \quad  t_{ij} \geq 0, \; \forall i, j,
\end{gather}
where $\langle \cdot, \cdot \rangle_F$ is the Frobenius inner product.
The dual formulation introduces variables $\alpha_i$ and $\beta_j$ that correspond to the marginal constraints:
\begin{gather}
\max_{\alpha, \beta} \left( \sum_{i=1}^{n} \frac{\alpha_i}{n} + \sum_{j=1}^{m} w_j \beta_j \right)\\
\textrm{subject to} \quad \alpha_i + \beta_j \leq c_{ij}, \forall i, j.
\end{gather}

Through strong duality \citep{boyd2004convex}, the optimal dual variables $\beta_j$ provide the subgradient of the objective with respect to $\mathbf{w}$. This elegant property allows us to efficiently optimize weights using projected subgradient descent, guiding mass toward locations that minimize transportation cost.

\paragraph{Position optimization with fixed weights}
With $\mathbf{w}$ fixed, the objective is quadratic in each $\tilde{\mathbf{x}}_j$; its (classical) Hessian is $\nabla^{2}_{\tilde{\mathbf{x}}_j} f = 2 w_j \mathbf I$. Performing one Newton step therefore amounts to
\begin{align}
  \tilde{\mathbf{x}}_j \leftarrow \tilde{\mathbf{x}}_j - \frac{1}{w_j}\sum_{i=1}^{n} t_{ij}(\tilde{\mathbf{x}}_j - \mathbf{x}_i).
\end{align}
Intuitively, this update pulls each synthetic point toward real data points based on the optimal transport plan, with the ``pull strength'' weighted by the transport allocation. Points with higher transport allocation exert stronger influence on the synthetic positions.

By alternating between these two optimization steps, we converge to a local optimum that represents the Wasserstein barycenter of the real data distribution. Remarkably, we find that even a small number of iterations produces high-quality synthetic data. Further details on this method are available in Appendix~\ref{sec: math_detail}.

\subsection{Barycenter Matching in the Feature Space}
\label{sec: engi}
Our above discussion shows that dataset distillation can be cast as the problem of finding the barycenter of the real data distribution, and there is an efficient approach for computing this barycenter. However, for high dimensional data such as images, it is beneficial to use some prior to learning synthetic images that encode meaningful information from the real dataset. 
Inspired by recent works \citep{yin2020dreaming, yin2023squeeze}, we use a pretrained classifier to embed the images into the feature space, in which we compute the Wasserstein barycenter to learn synthetic images. This subsection details our concrete algorithm design, which is illustrated in \cref{fig:wmdd-scheme}, and summarized in \cref{alg: wasserstein_barycenter}.

\begin{algorithm}[t]
    \caption{Wasserstein Metric-based Dataset Distillation (WMDD)}
    \label{alg: wasserstein_barycenter}
    
    \KwIn{Real dataset $\mathcal{T} = \{\mathbf{x}_{k, i}\}_{i=1, \ldots, n_k}^{k=1, \ldots, g}$, teacher model $f$ with feature extractor $f_e$ (before the linear classifier), number of iterations $K$}
    
    Train model $f$ on $\mathcal{T}$\;
    
    \For{each class $k$}{
        \For{each sample $i$}{
            Perform forward pass: $f(\mathbf{x}_{k, i})$\;
            Store feature: $f_e(\mathbf{x}_{k, i})$\;
        }
        Compute $\text{BN}_{k, l}^{\text{mean}}$, $\text{BN}_{k, l}^{\text{var}}$\;
    }
    
    \For{each class $k$}{
        $\{\mathbf{b}_{k, j}\}_{j=1, \ldots, m_k}, \{\mathbf{w}_{k, j}\}_{j=1, \ldots, m_k}
        \gets \operatorname{barycenter}\left(\{f_e(\mathbf{x}_{k, i})\}_{i=1, \ldots, n_k}\right)$, according to Algorithm~\ref{alg: bary_learning} (in Appendix~\ref{sec: alg_detail}) with $K$ iterations\;
        Optimize $\{\tilde{\mathbf{x}}_{k, j}\}_{j=1, \ldots, m_k}$ according to Eq.~\ref{eq: total_loss}\;
    }
    \KwOut{Synthetic dataset $\mathcal{S} $ with positions $\{\tilde{\mathbf{x}}_{k, j}\}_{j=1, \ldots, m_k}^{k=1, \ldots, g}$ and weights $\{\mathbf{w}_{k, j}\}_{j=1, \ldots, m_k}^{k=1, \ldots, g}$.}
\end{algorithm}

Suppose the real dataset $\mathcal{T}$ has $g$ classes, with $n_k$ images for class $k$ (hence $n=\sum_{k=1}^{g} n_k$). Let us re-index the samples by classes and denote the training set as $\mathcal{T} = \{\mathbf{x}_{k, i}\}_{i=1, \ldots, n_k}^{k=1, \ldots, g}$. Suppose that we want to distill $m_k$ images for class $k$. Denote the synthetic set $\mathcal{S} = \{\tilde{\mathbf{x}}_{k, j}\}_{j=1, \ldots, m_k}^{k=1, \ldots, g}$, where $m_k \ll n_k$ for all $k$.

First, we employ the pretrained model to extract features for all samples within each class in the original dataset $\mathcal{T}$. More specifically, we use the pretrained model $f$ to obtain the feature set $\{f_e(\mathbf{x}_{k, i})\}_{i=1, \ldots, n_k}$ for each class $k$, where $f_e(\cdot)$ returns the representation immediately before the linear classifier.

Next, we compute the Wasserstein barycenter for each feature set computed in the previous step. We treat the feature set for each class as an empirical distribution, and adapt the algorithm in \citep{cuturi2014fast} to compute the free support barycenters with $m_k$ points for class $k$, denoted as $\{\mathbf{b}_{k, j}\}_{j=1, \ldots, m_k}$, and the associated weights $\{\mathbf{w}_{k, j}\}_{j=1, \ldots, m_k}$, which are used to weight the synthetic images. 

Then, in the main distillation process, 
we use iterative gradient descent to learn the positions of synthetic images by jointly considering two objectives. We match the features of the synthetic images with the corresponding data points in the learned barycenter:
\begin{align}
\label{eq: bary_loss}
\mathcal{L}_{\text{feature}}(\tilde{\mathbf{X}}) &= \sum_{k=1}^g \sum_{j=1}^{m_k}  \| f_e (\tilde{\mathbf{x}}_{k, j}) - \mathbf{b}_{k, j} \|_2^2,
\end{align}
where $f_e(\cdot)$ is the function to compute features of the last layer. 

\begin{table*}[!htbp]
\centering
\setlength{\tabcolsep}{1.3mm}
\footnotesize
\renewcommand{\arraystretch}{1.15}
\begin{tabular}{@{}lcccc@{\hspace{4mm}}cccc@{\hspace{4mm}}cccc@{}}
\toprule
\multirow{2}{*}{Methods} & \multicolumn{4}{c}{\hspace{-1.8mm}ImageNette} & \multicolumn{4}{c}{\hspace{-4.8mm}Tiny ImageNet} & \multicolumn{4}{c}{\hspace{-1.3mm}ImageNet-1K} \\ 
\cmidrule(l{1.8mm}r{3.8mm}){2-5} \cmidrule(l{-0.4mm}r{3.9mm}){6-9} \cmidrule(l{-0.6mm}r{1mm}){10-13} 
                         & 1    & 10   & 50   & 100  & 1    & 10   & 50   & 100  & 1    & 10   & 50   & 100  \\ 
\hline
Random \citep{zhao2023dataset}          & 23.5 {\tiny $\pm$ 4.8} & 47.7 {\tiny $\pm$ 2.4} & {-}                    & {-}                     
                                        & 1.5 {\tiny $\pm$ 0.1}  & 6.0 {\tiny $\pm$ 0.8}  & 16.8 {\tiny $\pm$ 1.8} & {-}                     
                                        & 0.5 {\tiny $\pm$ 0.1}  & 3.6 {\tiny $\pm$ 0.1}  & 15.3 {\tiny $\pm$ 2.3} & {-}                     \\
                                        
DM \citep{zhao2023dataset}              & 32.8 {\tiny $\pm$ 0.5} & 58.1 {\tiny $\pm$ 0.3} & {-}                     & {-}                     
                                        & 3.9 {\tiny $\pm$ 0.2}  & 12.9 {\tiny $\pm$ 0.4} & 24.1 {\tiny $\pm$ 0.3}  & {-}                     
                                        & 1.5 {\tiny $\pm$ 0.1}  & {-}                    & {-}                     & {-}                     \\
                                        
MTT \citep{cazenavette2022distillation} & \textbf{47.7} {\tiny $\pm$ 0.9} & 63.0 {\tiny $\pm$ 1.3} & {-}                    & {-}                     
                                        & \textbf{8.8} {\tiny $\pm$ 0.3}  & 23.2 {\tiny $\pm$ 0.2} & 28.0 {\tiny $\pm$ 0.3} & {-}            
                                        & {-}                             & {-}                    & {-}                    & {-}            \\
                                        
DataDAM \citep{sajedi2022datadam}       & 34.7 {\tiny $\pm$ 0.9} & 59.4 {\tiny $\pm$ 0.4} & {-}                    & {-}                     
                                        & 8.3 {\tiny $\pm$ 0.4}  & 18.7 {\tiny $\pm$ 0.3} & 28.7 {\tiny $\pm$ 0.3} & {-}                     
                                        & 2.0 {\tiny $\pm$ 0.1}  & 6.3 {\tiny $\pm$ 0.0}  & 15.5 {\tiny $\pm$ 0.2} & {-}                     \\
                                        
SRe$^2$L \citep{yin2023squeeze}         & {20.6\textsuperscript{†} {\tiny $\pm$ 0.3}} & {54.2\textsuperscript{†} {\tiny $\pm$ 0.4}} & {80.4\textsuperscript{†} {\tiny $\pm$ 0.4}} & {85.9\textsuperscript{†}{\tiny $\pm$ 0.2}} 
                                        & {-}                                         & {-}                                         & 41.1 {\tiny $\pm$ 0.4}                      & 49.7 {\tiny $\pm$ 0.3} 
                                        & {-}                                         & 21.3 {\tiny $\pm$ 0.6}                      & 46.8 {\tiny $\pm$ 0.2}                      & 52.8 {\tiny $\pm$ 0.4} \\
                                        
CDA\textsuperscript{‡} \citep{yin2023dataset} & {-}               & {-}                     & {-}                              & {-}                     
                                              & {-}               & {-}                     & 48.7 \phantom{{\tiny $\pm$ 0.3}} & 53.2 \phantom{{\tiny $\pm$ 0.3}} 
                                              & {-}               & {-}                     & 53.5 \phantom{{\tiny $\pm$ 0.3}} & 58.0 \phantom{{\tiny $\pm$ 0.3}} \\
                                               
G-VBSM \citep{shao2024generalized}      & {-}                     & {-}                     & {-}                     & {-}                     
                                        & {-}                     & {-}                     & 47.6 {\tiny $\pm$ 0.3}  & 51.0 {\tiny $\pm$ 0.4} 
                                        & {-}                     & 31.4 {\tiny $\pm$ 0.5}  & 51.8 {\tiny $\pm$ 0.4}  & 55.7 {\tiny $\pm$ 0.4} \\
                                        
SCDD \citep{zhou2024self}               & {-}                     & {-}                     & {-}                     & {-}                     
                                        & {-}                     & 31.6 {\tiny $\pm$ 0.1}  & 45.9 {\tiny $\pm$ 0.2}  & {-}                     
                                        & {-}                     & 32.1 {\tiny $\pm$ 0.2}  & 53.1 {\tiny $\pm$ 0.1}  & 57.9 {\tiny $\pm$ 0.1} \\
\rowcolor{green!10}[6.4pt]
\textbf{WMDD}                           & 40.2 {\tiny $\pm$ 0.6}         & \textbf{64.8} {\tiny $\pm$ 0.4} & \textbf{83.5} {\tiny $\pm$ 0.3} & \textbf{87.1} {\tiny $\pm$ 0.3} 
                                        & 7.6 {\tiny $\pm$ 0.2}          & \textbf{41.8} {\tiny $\pm$ 0.1} & \textbf{59.4} {\tiny $\pm$ 0.5} & \textbf{61.0} {\tiny $\pm$ 0.3} 
                                        & \textbf{3.2} {\tiny $\pm$ 0.3} & \textbf{38.2} {\tiny $\pm$ 0.2} & \textbf{57.6} {\tiny $\pm$ 0.5} & \textbf{60.7} {\tiny $\pm$ 0.2} \\
\bottomrule
\end{tabular}
\caption{Performance comparison of various dataset distillation methods on different datasets. We used the reported results for baseline methods when available. We replicated the result of SRe$^2$L on the  ImageNette dataset, marked by \textsuperscript{†}. Results of CDA did not include error bars, and the row is marked by \textsuperscript{‡}.}
\label{tab: performance_main}
\end{table*}

To further leverage the capability of the pretrained model in aligning the distributions, previous DD works \citep{yin2020dreaming, yin2023squeeze} have used BatchNorm statistics of the real data to regularize synthetic images. However, the gradient on each synthetic sample for optimizing global BN alignment in a batch of mixed classes may not synergize well with the gradient on the same sample for matching its class-specific objective like the CE loss. Intuitively, the BN statistics within different data classes may vary, and simply encouraging alignment of global BN statistics does not provide enough information about how synthetic samples from different classes should contribute differently to the global BN statistics, potentially leading to suboptimal distillation quality.

Thus, to better capture the intra-class data distribution, we propose the Per-Class BatchNorm (PCBN) regularization method, using BatchNorm statistics of the real data \emph{within each class separately} to regularize synthetic data. 
While the method might sound conceptually similar to previous variants of BatchNorm for addressing feature distribution heterogeneity \cite{gong2022sandwich} and long-tail problems \cite{cheng2022compound} in different areas of computer vision, it is fundamentally different in technical design.
Specifically, we regularize synthetic images with
\begin{equation}
\label{eq: bn_reg}
\resizebox{\linewidth}{!}{$\displaystyle
\begin{aligned}
\mathcal{L}_{\text{BN}}(\tilde{\mathbf{X}})=\sum_{k=1}^{g}\sum_{l=1}^{L}\Bigl(&\,\| \mathcal{A}_{\text{mean}}(\{f_l(\tilde{\mathbf{x}}_{k,j})\}_{j=1}^{m_k},\{w_{k,j}\}_{j=1}^{m_k})-\text{BN}_{k,l}^{\text{mean}}\|_2^2\\
&+\,\| \mathcal{A}_{\text{var}}(\{f_l(\tilde{\mathbf{x}}_{k,j})\}_{j=1}^{m_k},\{w_{k,j}\}_{j=1}^{m_k})-\text{BN}_{k,l}^{\text{var}}\|_2^2 \Bigr).
\end{aligned}$}
\end{equation}

\noindent Here, $L$ is the number of BatchNorm layers, and $f_l(\cdot)$ is the function that computes the feature map that feeds the $l$-th BatchNorm layer. ${BN}_{k,l}^{\text{mean}}$ and ${BN}_{k,l}^{\text{var}}$ denote the per-channel mean and variance of class~$k$, obtained from one pass over the real data. \vspace{2pt}
The weighted aggregate operators $\mathcal{A}_{\text{mean}}$ and $\mathcal{A}_{\text{var}}$ compute statistics of synthetic samples while respecting the optimal transport weights. For feature tensor $\mathbf{F}$ with spatial dimensions $H \times U$ (height and width), these operators compute channel-wise statistics:
\begin{equation}
\label{eq:aggregate_ops}
\resizebox{1.02\linewidth}{!}{$\displaystyle
\begin{aligned}
\mathcal{A}_{\text{mean}}(\mathbf{F},\mathbf{w})_c &:= \frac{1}{H U\sum_{j=1}^{m_k} w_{k,j}} \sum_{j=1}^{m_k} w_{k,j} \sum_{h=1}^{H}\sum_{u=1}^{U} F_{j,c,h,u},\\[4pt]
\mathcal{A}_{\text{var}}(\mathbf{F},\mathbf{w})_c &:= \frac{1}{H U\sum_{j=1}^{m_k} w_{k,j}} \sum_{j=1}^{m_k} w_{k,j} \sum_{h=1}^{H}\sum_{u=1}^{U}\bigl(F_{j,c,h,u}-\mathcal{A}_{\text{mean}}(\mathbf{F},\mathbf{w})_c\bigr)^{2}.
\end{aligned}$}
\end{equation}
Here, $F_{j,c,h,u}$ denotes the activation at position $(h,u)$ in channel $c$ for synthetic sample $j$. Each expression computes statistics for channel $c$; concatenating across all channels yields the complete mean and variance vectors.

Combining these objectives above, we employ the below loss function for learning the synthetic data:
\begin{equation}
\label{eq: total_loss}
\mathcal{L}(\tilde{\mathbf{X}}) = \mathcal{L}_{\text{feature}}(\tilde{\mathbf{X}}) + \lambda \mathcal{L}_{\text{BN}}(\tilde{\mathbf{X}}),
\end{equation}
where $\lambda$ is a regularization coefficient. The synthetic set $\mathcal{S}$ therefore comprises the positions $\tilde{\mathbf{X}}$ and their associated weights $\{w_{k,j}\}_{j=1}^{m_k}$, which are used in the FKD stage following previous DD works \cite{yin2020dreaming, yin2023squeeze, shao2024generalized}.
     
\section{Experiments}
\label{sec:experiment}
\subsection{Experiment Setup}
We systematically evaluated our method on three high-resolution datasets: ImageNette \citep{howard2019imagenette}, Tiny ImageNet \citep{le2015tiny}, and ImageNet-1K \citep{deng2009imagenet}. We tested synthetic image budgets of 1, 10, 50, and 100 images per class (IPC).
For each dataset, we trained a ResNet-18 model \citep{he2016deep} on the real training set, distilled the dataset using our method, then trained a ResNet-18 model from scratch on the synthetic data. We measured performance using the top-1 accuracy of the trained model on the validation set. Results report the mean and standard deviation from 3 repeated runs.
Our barycenter algorithm implementation used the Python Optimal Transport library \citep{flamary2021pot}. We maintained most hyperparameter settings from \citep{yin2023squeeze} but adjusted our loss terms' regularization coefficient $\lambda$.
For barycenter computation (\Cref{alg: wasserstein_barycenter}), we found $K=10$ iterations sufficient for high-performance synthetic data generation. Increasing $K$ yielded only marginal improvements, so we kept $K=10$ to balance efficiency and performance. We provide full implementation details in Appendix \ref{sec: impl_detail}.

\subsection{Comparison with Other Methods}
With this experimental setup, we now evaluate how our Wasserstein metric-based approach performs against existing dataset distillation methods.

\definecolor{myblue}{HTML}{5DA5DA}
\definecolor{myorange}{HTML}{FAA43A}
\definecolor{mygreen}{HTML}{60BD68}
\definecolor{myred}{HTML}{FF6347} 
\definecolor{mypurple}{HTML}{B276B2}

We compared our method against several baselines and recent strong dataset distillation (DD) approaches, including distribution matching-based methods like DataDAM \citep{sajedi2022datadam}, SRe$^{2}$L \citep{yin2023squeeze}, CDA \citep{yin2023dataset}, G-VBSM \citep{shao2024generalized}, and SCDD \citep{zhou2024self}, selected for their scalability to large, high-resolution datasets. Table~\ref{tab: performance_main} presents our experimental results alongside reported results from these methods under identical settings.
Our method consistently achieved state-of-the-art performance in most settings across different datasets. Compared to MTT \citep{cazenavette2022distillation} and DataDAM \citep{sajedi2022datadam}, which show good performance in fewer IPC settings, the performance of our method increases more rapidly with the number of synthetic images. 
Notably, in the 100 IPC setting, our method achieved top-1 accuracies of 87.1\%, 61.0\%, and 60.7\% across the three datasets, respectively. These results approach those of pretrained classifiers (89.9\%, 63.5\%, and 63.1\%) trained on full datasets. This superior performance highlights the effectiveness and robustness of our approach in achieving higher accuracy across different datasets.

\subsection{Cross-architecture Generalization}
Beyond achieving strong performance on the distillation architecture, a critical test for any dataset distillation method 
is how well the synthetic data generalizes to different model architectures \citep{cazenavette2023glad, lei2022comprehensive}. 
For this aim, we conducted experiments training various randomly initialized models on synthetic data generated 
via our ResNet-18-based method. To prevent overfitting on the small synthetic data while ensuring fair comparison, 
we held out 20\% of the distilled data as validation set to find the best training epoch for each experiment. 
We report the performance of different evaluation models, ResNet-18, ResNet-50, ResNet-101 \citep{he2016deep}, ViT-Tiny, and ViT-Small 
\citep{dosovitskiy2020image}, in the 50 IPC setting on ImageNet-1K. 
The results in Table \ref{tab:cross_architecture_table} show that our method demonstrates stronger cross-architecture transfer than previous methods. 
Our synthetic data generalizes well across the ResNet family, where the performance increases with the model capacity. 
The performance on the vision transformers is relatively lower, probably due to their data-hungry property.

\begin{table}[h]
\centering
\small
\renewcommand{\arraystretch}{1.1}
\begin{tabular}{c@{\hspace{6.2pt}}ccccc}
\toprule
Method      &  Res18         &  Res50         &  Res101        & ViT-T          &  ViT-S \\ 
\midrule
SRe$^2$L    &  48.02         &  55.61         &  60.86         & 16.56          &  15.75 \\
CDA         &  54.43         &  60.79         &  61.74         & 31.22          &  32.97 \\
G-VBSM      &  52.28         &  59.08         &  59.30         & 30.30          &  30.83 \\
\rowcolor{green!10}
WMDD (Ours) & \textbf{57.83} & \textbf{61.22} & \textbf{62.57} & \textbf{34.25} & \textbf{34.87} \\
\bottomrule
\end{tabular}
\caption{\small Cross-architecture generalization performance on ImageNet-1K in 50 IPC setting. We used ResNet-18 for distillation and different architectures for evaluation: ResNet-\{18,50,101\}, ViT-Tiny and ViT-Small with a patch size of 16.}
\label{tab:cross_architecture_table}
\end{table}

\subsection{Efficiency Analysis}
\label{sec:efficiency}
Having demonstrated the effectiveness of our approach, we now examine its computational efficiency—a crucial factor for 
practical deployment. To evaluate the time and memory efficiency of our method, we measured the time used per iteration, 
total computation time, and the peak GPU consumption of our method with a 3090 GPU on ImageNette in the 1 IPC setting and 
compared these metrics among several different methods. The results are shown in Table~\ref{tab:time}. As the Wasserstein 
barycenter can be computed efficiently, our method only brings minimal additional computation time compared with most 
efficient methods such as \citep{yin2023squeeze}. This makes it possible to preserve the efficiency benefits of the 
distribution-based method while reaching strong performance.

\begin{table}[htbp!]
\centering
\setlength{\tabcolsep}{5pt} 
\small
\renewcommand{\arraystretch}{1.1}
\begin{tabular}{cccc}
\toprule
Method   & Time/iter (s)                           & Peak vRAM (GB) & Total time (s) \\
\midrule
DC       & 2.154 {\tiny $\pm$ 0.104}               & 11.90          & 6348.17 \\
DM       & 1.965 {\tiny $\pm$ 0.055}               & 9.93           & 4018.17 \\
SRe$^2$L & 0.015 {\tiny $\pm$ 0.029}               & \textbf{1.14}  & \textbf{194.90} \\
\rowcolor{green!10}
\textbf{WMDD} & \textbf{0.013 {\tiny $\pm$ 0.001}} & 1.22           & 207.53 \\
\bottomrule
\end{tabular}
\caption{Distillation time and GPU memory usage on ImageNette using a single GPU (RTX-3090) for all methods. `Time/iter' indicates the time to update 1 synthetic image per class with a single iteration. This duration is measured in consecutive 100 iterations, and the mean and standard deviation are reported. For a fair comparison, we keep the original image resolution and use the ResNet-18 model to distill 2,000 iterations for all methods.
}
\label{tab:time}
\end{table}

\subsection{Ablation Study}
\label{sec: ablation}
To understand the individual contributions of our key design choices, we conducted an ablation study examining which factors 
drive our method's improved performance. We examined two key factors: whether to use our Wasserstein barycenter loss 
(Eq.~\ref{eq: bary_loss}) or the cross-entropy loss \citep{yin2023squeeze, yin2023dataset} for feature matching; and whether 
to use standard BatchNorm statistics or our PCBN method for regularization.
We evaluated these factors across different datasets using the 10 IPC setting, with results shown in Table \ref{tab:ablation}.
As discussed in our method design (\Cref{sec: engi}), standard BN computes statistics from all-class samples, which does not 
synergize well with the class-specific matching objective, leading to mixed results with the Wasserstein loss. In contrast, our 
PCBN method significantly improves performance on all datasets by capturing intra-class distributions.
When properly paired with PCBN, our Wasserstein loss yields further significant gains across all datasets. 
As our WMDD method already achieves high performance (with our 100 IPC results approaching those of full dataset training), these consistent 
improvements confirm the effectiveness of our design choices.

\begin{table}[htbp!]
\centering
\setlength{\tabcolsep}{1pt}
\small
\renewcommand{\arraystretch}{1.1}
\label{Performance}
\begin{tabular}{M{1cm} M{0.8cm} M{2cm} M{2cm} M{2cm} }
\toprule
$\mathcal{L}_{\text{feature}}$ & $\mathcal{L}_{\text{reg}}$ & ImageNette & Tiny ImageNet & ImageNet-1K \\ 
\midrule
\rowcolor{green!10}
\textbf{Wass.} & \textbf{PCBN} & \textbf{64.7{\tiny $\pm$ 0.2}} & \textbf{41.8{\tiny $\pm$ 0.1}} & \textbf{38.1{\tiny $\pm$ 0.1}} \\
CE & PCBN                      & 63.5{\tiny $\pm$ 0.1}          & 41.0{\tiny $\pm$ 0.2}          & 36.4{\tiny $\pm$ 0.2} \\
Wass. & BN                     & 60.7{\tiny $\pm$ 0.2}          & 36.6{\tiny $\pm$ 0.1}          & 26.8{\tiny $\pm$ 0.3} \\
CE & BN                        & 54.2{\tiny $\pm$ 0.1}          & 38.0{\tiny $\pm$ 0.3}          & 35.9{\tiny $\pm$ 0.2} \\
\bottomrule
\end{tabular}
\caption{Ablation study on two variables: whether to use our Wasserstein (Wass.) loss or the cross-entropy (CE) loss in previous DD works \cite{yin2023squeeze, yin2023dataset} for feature matching ($\mathcal{L}_{\text{feature}}$), and whether to use standard BatchNorm (BN) or our PCBN method for regularization ($\mathcal{L}_{\text{reg}}$). We report the mean and standard error of performance on 5 repetitive runs.}
\label{tab:ablation}
\end{table} 

Additionally, we find that directly replacing the Wasserstein metric in our method with MMD results in near-random performance 
on Tiny-ImageNet and ImageNet-1K. This motivates a deeper analysis of different distribution metrics, which we provide below.

\subsection{Comparison with Alternative Metrics}
\label{sec: alternative_metrics}
\begin{figure}[h]
    \small 
    \centering
    \includegraphics[width=\linewidth]{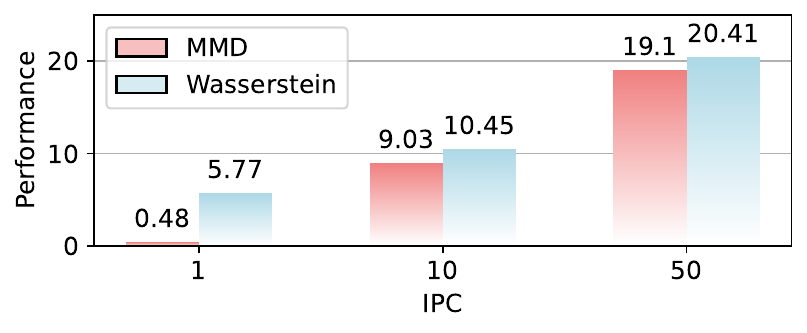}
    \caption{Performance comparison of MMD distance vs. the Wasserstein distance. The evaluation model is ResNet18.}
    \label{fig:compare_mmd}
\end{figure}

\paragraph{The MMD Metric}
\Cref{tab: performance_main} shows that our method using the Wasserstein metric outperforms all previous DD methods, including 
MMD-based methods such as \cite{zhao2023dataset}. A more direct comparison between the two distribution metrics is tricky, 
because existing MMD-based methods require feature spaces from dozens of randomly initialized models, which is incompatible 
with our algorithm using a single pretrained model. Simply replacing the Wasserstein metric in our method with MMD results in 
near-random performance. To try to make a fair comparison, we removed engineering tricks from DD methods using both metrics 
and evaluated their vanilla versions on Tiny-ImageNet.
Specifically, we compared our method with a seminal MMD-based method \citep{zhao2023dataset} on Tiny-ImageNet, and removed all 
engineering tricks including fancy augmentations (e.g., rotation, color jitter, and mixup) used in both methods and the FKD 
\citep{shen2022fast} used in our method. According to the result in Figure \ref{fig:compare_mmd}, the Wasserstein metric 
yields better synthetic data in all settings. In 1 IPC setting, the MMD metric yields random performance, likely due to 
empirical approximation errors and its focus on feature means rather than their geometric properties. 
In Appendix~\ref{sec: theory}, we provide a possible theoretical explanation for the superior performance of the Wasserstein 
metric by combining error bound analysis with the practicality of existing MMD-based methods.

\paragraph{The Sliced Wasserstein Distance}
Beyond MMD, we also examined the Sliced Wasserstein (SW) distance \cite{nguyen2023energy}, which has shown promise in reducing 
computational cost while retaining key aspects of Wasserstein geometry. In Table~\ref{tab: sliced_wass}, we compare our Wasserstein barycenters to those 
computed with SW and show that the latter achieves comparable accuracy with a modest increase in speed. However, our full 
barycenter computation is already highly efficient, accounting for only a small fraction of the overall runtime.
\begin{table}[h]
\centering
\setlength{\tabcolsep}{7.5pt}
\resizebox{0.45\textwidth}{!}{
\scriptsize
\begin{tabular}{@{}l|ccc|ccc@{}}
\toprule
\scriptsize Method & \multicolumn{3}{c|}{\scriptsize Accuracy (\%)} & \multicolumn{3}{c}{\scriptsize Time (hour)}  \\
\cmidrule(lr){2-4} \cmidrule(l){5-7}
\scriptsize IPC & \scriptsize 1 & \scriptsize 10 & \scriptsize 50 & \scriptsize 1 & \scriptsize 10 & \scriptsize 50 \\
\midrule
\rowcolor{green!10}
WMDD (Ours)     & \textbf{7.6}  & \textbf{41.8}  & \textbf{59.4}  & 0.71          & 2.30 & 5.27                     \\
Sliced Wass.    & 7.4           & 41.1           & 58.3           & \textbf{0.68} & \textbf{2.23}  & \textbf{5.16}  \\
\bottomrule
\end{tabular}}
\caption{Performance and efficiency comparison with Sliced Wass. Distance on Tiny-ImageNet.}
\label{tab: sliced_wass}
\vspace{-3pt}
\end{table}



\subsection{Hyperparameter Sensitivity}
The robustness of our method to hyperparameter choices is important for practical applications. We analyze sensitivity to key 
hyperparameters below.

\begin{figure}[h]
    \centering
    \begin{subfigure}[b]{0.49\textwidth}
        \centering
        \begin{tikzpicture}
        \begin{axis}[
            xlabel={\textit{$\lambda$} (logarithmic scale, base 10)},
            xlabel style={xshift=-15pt},
            ylabel={Performance Metric},
            ylabel style={yshift=-15pt},
            ymin=0,
            xmode=log,
            log basis x={10},
            grid=both,
            legend pos=south east,
            legend style={legend columns=3},
            mark size=2pt,
            width=\textwidth, 
            height=0.5\textwidth, 
            font=\scriptsize
        ]
        \addplot[
            color=myblue,
            mark=o,
            line width=1.5pt
        ]
        coordinates {
            (0.1,12.27) (0.3,22.14) (1,26.61) (3,31.98) (10,35.55) (30,35.76) (100,37.11) (300,37.68) (1000,37.53)
        };
        \addlegendentry{ImageNet}
        
        \addplot[
            color=myred,
            mark=square,
            line width=1.5pt
        ]
        coordinates {
            (0.1,58.27) (0.3,58.91) (1,58.85) (3,63.54) (10,63.23) (30,63.39) (100,64.31) (300,64.15) (1000,64.28)
        };
        \addlegendentry{Imagenette}
        
        \addplot[
            color=mygreen,
            mark=triangle,
            line width=1.5pt
        ]
        coordinates {
            (0.1,30.14) (0.3,36.46) (1,38.23) (3,36.02) (10,37.26) (30,38.88) (100,39.96) (300,41.98) (1000,40.76)
        };
        \addlegendentry{Tiny ImageNet}
        
        \end{axis}
        \end{tikzpicture}
        \caption{Effect of \textit{$\lambda$} on WMDD performance on the three datasets.}
        \label{fig:dataset-performance}
    \end{subfigure}
    \\
    \begin{subfigure}[b]{0.48\textwidth}
        \centering
        \begin{adjustbox}{width=\linewidth,center}
        \setlength{\fboxsep}{0pt}
        \begin{tabular}{@{}c@{\hspace{1pt}}c@{\hspace{2pt}}c@{\hspace{1pt}}c@{}}
            \fbox{\includegraphics[width=0.17\linewidth]{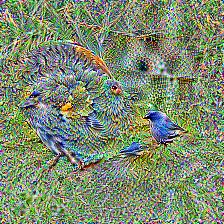}} &
            \fbox{\includegraphics[width=0.17\linewidth]{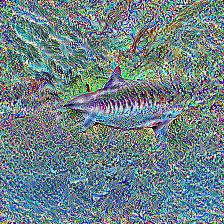}} &
            \fbox{\includegraphics[width=0.17\linewidth]{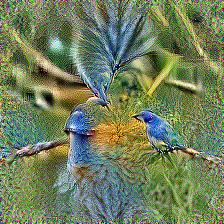}} &
            \fbox{\includegraphics[width=0.17\linewidth]{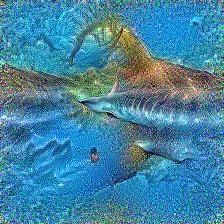}}
            \\
            
            \multicolumn{2}{c}{\tiny $\lambda=0.1$} & \multicolumn{2}{c}{\tiny $\lambda=1$} \\
            
            \fbox{\includegraphics[width=0.17\linewidth]{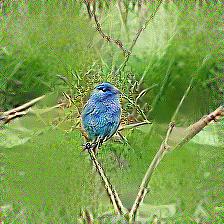}} &
            \fbox{\includegraphics[width=0.17\linewidth]{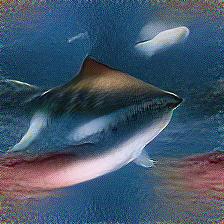}} &
            \fbox{\includegraphics[width=0.17\linewidth]{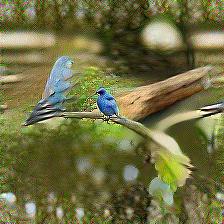}} &
            \fbox{\includegraphics[width=0.17\linewidth]{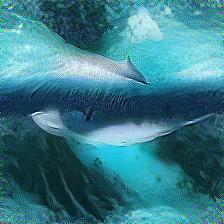}}
            \\
            \multicolumn{2}{c}{\tiny $\lambda=100$} & \multicolumn{2}{c}{\tiny $\lambda=1000$} \\
        \end{tabular}
        \end{adjustbox}
        \caption{Visualization of synthetic images from Imagenet-1K of classes indigo bird (left) and tiger shark (right), with different $\lambda$.}
        \label{fig:Synthetic_Image_ImageNet_compare}
    \end{subfigure}
    \caption{Effect of regularization strength \textit{$\lambda$} on our method.}
    \vspace{-10pt}
\end{figure}

\paragraph{Regularization Strength}
To analyze how the regularization term affects our method (Eq.~\ref{eq: total_loss}), we tested $\lambda$ values ranging from 
$10^{-1}$ to $10^3$ and evaluated performance on three datasets in 10 IPC setting. Figure \ref{fig:dataset-performance} 
shows that small $\lambda$ result in lower performance across all datasets. Performance improves as $\lambda$ increases, 
stabilizing around a threshold of approximately $10.0$. This demonstrates that while regularization enhances dataset quality, 
our method remains robust to specific $\lambda$ values.
Figure \ref{fig:Synthetic_Image_ImageNet_compare} illustrates the regularization effect on synthetic images of the same class. 
When $\lambda$ is too small, synthetic images exhibit high-frequency components, suggesting overfitting to model weights and 
architecture. In contrast, sufficiently large $\lambda$ values produce synthetic images that 
better align with human perception.

\begin{figure*}[t]
    \centering
    \begin{minipage}[t]{0.47\linewidth}
        \centering
        \includegraphics[width=\linewidth]{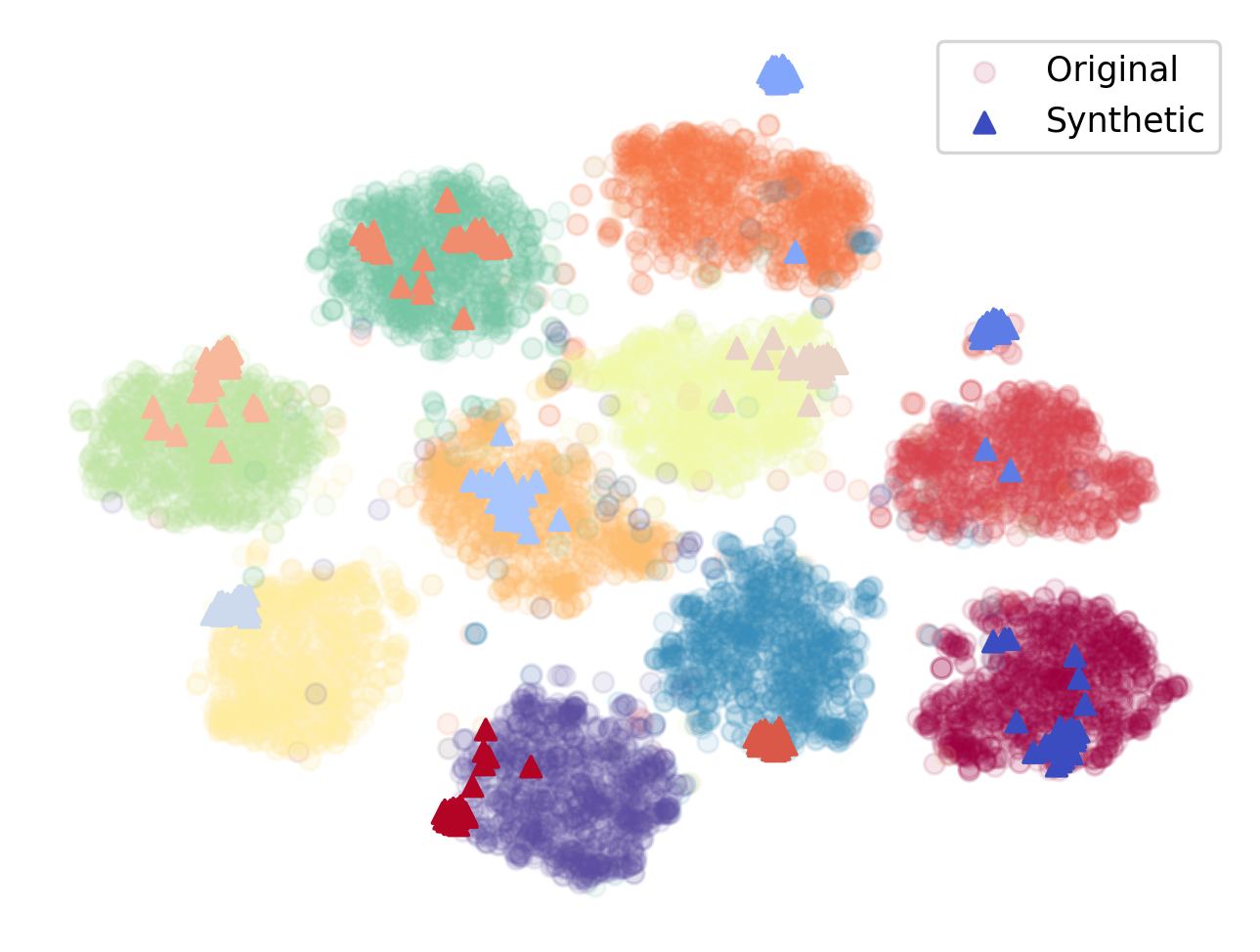}
        \label{fig:baseline_distribution}
    \end{minipage}
    \hfill
    \begin{minipage}[t]{0.47\linewidth}
        \centering
        \includegraphics[width=\linewidth]{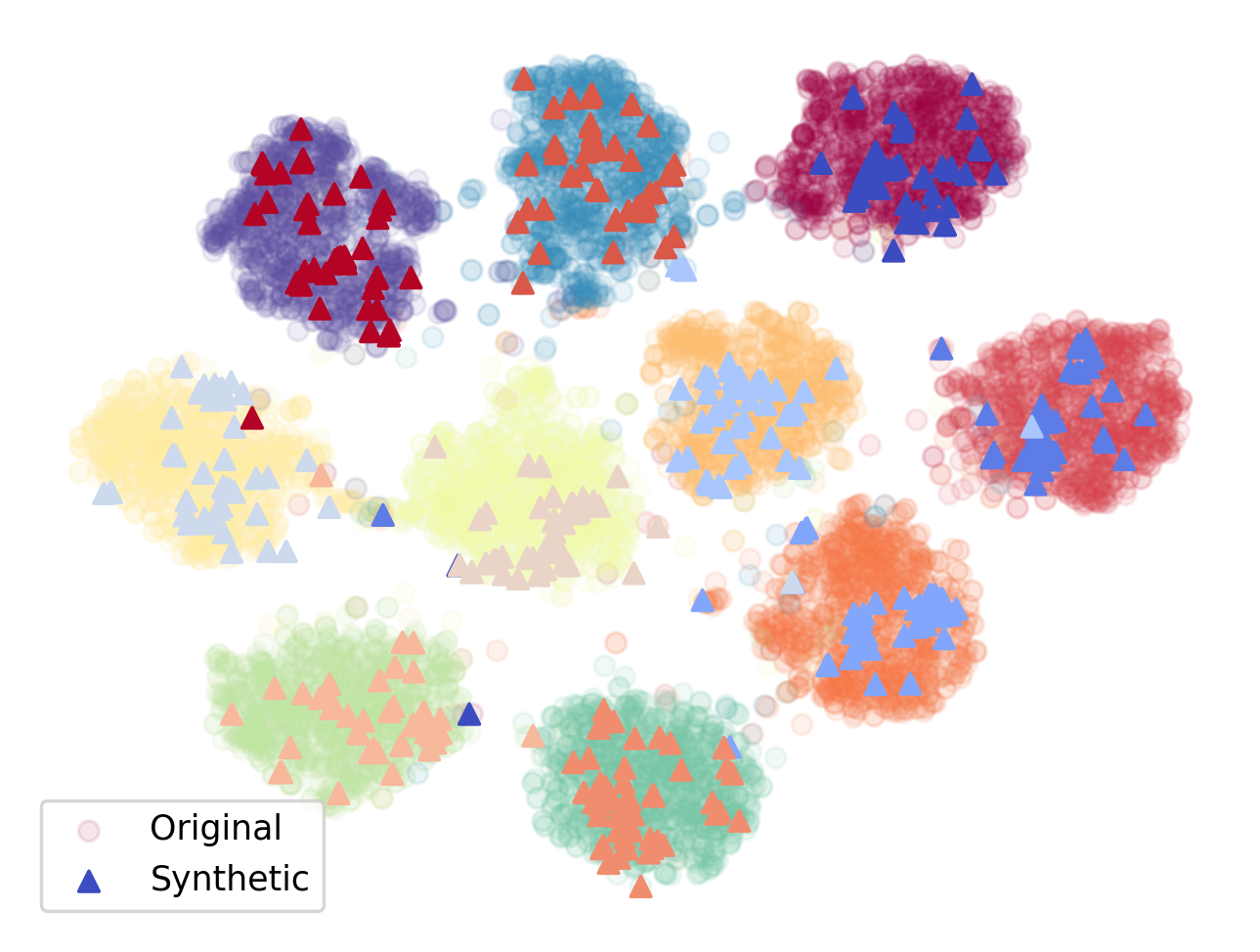}
        \label{fig:ours_distribution}
    \end{minipage}
    \caption{Distribution visualization of ImageNette. The dots present the original dataset's distribution using the model's latent space (e.g., ResNet-101), and the triangles are distilled images. Left: data distilled by SRe$^2$L; Right: data distilled by our method.}
    \label{fig:distribution_results}
\end{figure*}

\paragraph{Features from Different Layers}
Beyond regularization strength, we also examined which network layer provides the most effective features for our Wasserstein 
barycenter computation. Table \ref{tab:diff_layer} shows the performance with features from different layers of ResNet-18 on 
Tiny-ImageNet. 
The accuracy increases and then stabilizes by Layer 16, indicating WMDD leverages high-level, abstract 
representations.
\vspace{-6pt}
\begin{table}[h]
\centering
\small
\begin{tabular}{@{}c|cccccccc@{}}
\toprule
Layer    & 5   & 10   & 15   & 16   & 17   & 18 \\
\midrule
Acc (\%) & 2.4 & 11.3 & 37.6 & 41.1 & 41.6 & \textbf{41.8} \\
\bottomrule
\end{tabular}
    \caption{Performance of WMDD using features from different layers of the backbone.}
    \label{tab:diff_layer}
\end{table}

\subsection{Feature Embedding Distribution}\label{sec: tsne}
To provide intuitive insight into why our method achieves superior performance, we visualize how our synthetic data are 
distributed relative to the real data in feature space. 
We train a model from scratch on a mixture of both data to map the real and synthetic data into the same feature space. 
Then we extract their last-layer features and use the t-SNE \citep{van2008visualizing} method to visualize their 
distributions on a 2D plane. 
For comparison, we conduct this process for the synthetic data obtained using our method and the SRe$^2$L 
\citep{yin2023squeeze} method as a baseline. Figure \ref{fig:distribution_results} shows the result. 
In the synthetic images learned by SRe$^2$L, synthetic images within the same class tend to collapse, 
and those from different classes tend to be far apart. 
This is probably a result of the cross-entropy loss they used, 
which optimizes the synthetic images toward maximal output probability from the pre-trained model. 
In contrast, our utilization of the Wasserstein metric enables synthetic images to better represent the distribution of real 
data, maintaining both intra-class diversity and inter-class relationships that are crucial for effective model training.   
\section{Conclusion}
\label{sec:conclusion}

This work introduces a new dataset distillation approach leveraging Wasserstein metrics, grounded in optimal transport theory, to achieve more precise distribution matching. 
Our method learns synthetic datasets by matching the Wasserstein barycenter of the data distribution in the feature space of pretrained models, combined with a simple regularization technique to leverage the prior knowledge in these models. 
Through empirical testing, our approach has demonstrated impressive performance across a variety of benchmarks,
highlighting its reliability and practical applicability in diverse scenarios. Findings from our controlled experiments corroborate the utility of Wasserstein metrics for capturing the essence of data distributions. 
Future work will aim to explore the integration of advanced metrics with generative methods, aligning with the broader goal of advancing data efficiency in computer vision.

\clearpage
\newpage
{
    \small
    \bibliographystyle{ieeenat_fullname}
    \bibliography{main}
}

\clearpage
\newpage
\onecolumn
\appendix
\clearpage
\setcounter{section}{0} 

\maketitlesupplementary
\renewcommand{\thesection}{\Alph{section}}

The supplementary material is structured as follows:
\begin{itemize}
    \item Appendix \ref{sec:social_impact} outlines the potential social impact of our work;
    \item Appendix \ref{sec: theory} provides a possible theoretical explanation why the Wasserstein metric shows superior performance to MMD in our experiments;
    \item Appendix \ref{sec: math_detail} discusses the method for Wasserstein barycenter computation in more detail;
    \item Appendix \ref{sec: alg_detail} presents our algorithmic framework;
    \item Appendix \ref{sec: impl_detail} presents our implementation details;
    \item Appendix \ref{sec: variety} discusses the increased variety in our synthetic images;
    \item Appendix \ref{sec: visualizations} provides more visualization examples.
\end{itemize}

\section{Discussion on Potential Social Impact}
\label{sec:social_impact}

Our method, focused on accurately matching data distributions, inherently reflects existing biases in the source datasets, potentially leading to automated decisions that may not be completely fair. This situation underscores the importance of actively working to reduce bias in distilled datasets, a critical area for further investigation. Despite this, our technique significantly improves efficiency in model training by reducing data size, potentially lowering energy use and carbon emissions. This not only benefits the environment but also makes AI technologies more accessible to researchers with limited resources. While recognizing the concern of bias, the environmental advantages and the democratization of AI research our method offers are believed to have a greater positive impact.

\section{Theoretical Explanation on the Superior Performance of the Wasserstein Metric}\label{sec: theory}

In this section, we provide a possible theoretical explanation for the observed superior performance of the Wasserstein metric over the MMD metric in our experiments (shown in \cref{fig:compare_mmd} of the main paper).

It is important to note that the performance of dataset distillation (DD) methods depends largely on various factors in the algorithmic framework, such as the choice of neural networks or kernels \cite{shao2024generalized}, image sampling strategies, loss function design \cite{zhao2023improved}, and techniques like factorization \cite{liu2022dataset} and FKD \cite{shen2022fast}. Additionally, high-resolution datasets, which pose challenges to most existing DD methods, often necessitate trading some precision for computational feasibility in algorithm design. Consequently, we do not aim to assert that the Wasserstein metric is consistently superior as a statistical metric for distribution matching in DD, nor do we believe this to be the case. Instead, we provide a theoretical explanation for the observed superior performance of the Wasserstein metric by combining error bound analysis with practical considerations in DD algorithms, hoping to provide some insights into this phenomenon.

We consider two methods for measuring the discrepancy between the synthetic distribution $\mathbb{Q}$ and the real data distribution $\mathbb{P}$: the Wasserstein distance and the empirical Maximum Mean Discrepancy (MMD). Specifically, we focus on the Wasserstein-1 distance $W_1$, as it provides a meaningful and tractable metric in our context. 

\subsection{Setup and Notation}

Let $\mathcal{X} \subset \mathbb{R}^d$ denote the input space (assumed to be compact), and $\mathcal{Y} \subset \mathbb{R}$ the label space. Let $\mathbb{P}$ be the real data distribution over $\mathcal{X}$, and $\mathbb{Q}$ the synthetic data distribution over $\mathcal{X}$. Let $f: \mathcal{X} \to \mathcal{Y}$ be the labeling function. We consider a hypothesis class $\mathcal{H}$ of functions $h: \mathcal{X} \to \mathcal{Y}$. The loss function is $\ell: \mathcal{Y} \times \mathcal{Y} \to [0, \infty)$, and we denote the composite loss function as $g(x) = \ell(h(x), f(x))$.

\subsection{Assumptions}

We make the following assumptions:

\begin{itemize}[noitemsep,topsep=1pt, leftmargin=*]
    \item \textbf{A1.} The composite loss function $g(x)$ is Lipschitz continuous with respect to $x$, with Lipschitz constant $L$:
    \begin{equation}
    \label{eq:lipschitz_loss}
    |g(x) - g(x')| = |\ell(h(x), f(x)) - \ell(h(x'), f(x'))| \leq L \| x - x' \|.
    \end{equation}

    \item \textbf{A2.} The input space $\mathcal{X}$ is compact.

    \item \textbf{A3.} The kernel $k(x, x')$ used in MMD calculations is a characteristic kernel. That means, $\mathrm{MMD}_k(\mathbb{P}, \mathbb{Q}) = 0$ implies $\mathbb{P} = \mathbb{Q}$.

    \item \textbf{A4.} The composite loss function $g(x)$ lies in the Reproducing Kernel Hilbert Space (RKHS) $\mathcal{H}_k$ associated with the kernel $k$, with RKHS norm $\| g \|_{\mathcal{H}_k} < \infty$.
\end{itemize}

\subsection{Theoretical Analysis}

Our goal is to bound the difference in expected losses between the real and synthetic distributions:
\begin{equation}
\label{eq:risk_difference}
\left| \mathbb{E}_{x \sim \mathbb{P}} [ g(x) ] - \mathbb{E}_{x \sim \mathbb{Q}} [ g(x) ] \right|.
\end{equation}

\paragraph{Bounding Using Wasserstein Distance}

Under Assumption \textbf{A1}, the function $g(x)$ is Lipschitz continuous with constant $L$. By the definition of the Wasserstein-1 distance $W_1$:
\begin{equation}
W_1(\mathbb{P}, \mathbb{Q}) = \inf_{\gamma \in \Pi(\mathbb{P}, \mathbb{Q})} \mathbb{E}_{(x, x') \sim \gamma} [ \| x - x' \| ],
\end{equation}
where $\Pi(\mathbb{P}, \mathbb{Q})$ is the set of all couplings of $\mathbb{P}$ and $\mathbb{Q}$.

Using any coupling $\gamma \in \Pi(\mathbb{P}, \mathbb{Q})$, we have:
\begin{align}
\left| \mathbb{E}_{\mathbb{P}}[g(x)] - \mathbb{E}_{\mathbb{Q}}[g(x)] \right| &= \left| \int_{\mathcal{X}} g(x) \, d\mathbb{P}(x) - \int_{\mathcal{X}} g(x') \, d\mathbb{Q}(x') \right| \nonumber \\
&= \left| \int_{\mathcal{X} \times \mathcal{X}} \left( g(x) - g(x') \right) \, d\gamma(x, x') \right| \nonumber \\
&\leq \int_{\mathcal{X} \times \mathcal{X}} | g(x) - g(x') | \, d\gamma(x, x') \nonumber \\
&\leq L \int_{\mathcal{X} \times \mathcal{X}} \| x - x' \| \, d\gamma(x, x') \nonumber \\
&= L \mathbb{E}_{(x, x') \sim \gamma} [ \| x - x' \| ].
\end{align}

Since this holds for any coupling $\gamma$, it holds in particular for the optimal coupling that defines $W_1(\mathbb{P}, \mathbb{Q})$:
\begin{equation}
\label{eq: wasserstein_bound}
\left| \mathbb{E}_{\mathbb{P}}[g(x)] - \mathbb{E}_{\mathbb{Q}}[g(x)] \right| \leq L W_1(\mathbb{P}, \mathbb{Q}).
\end{equation}

This bound shows that minimizing the Wasserstein-1 distance $W_1(\mathbb{P}, \mathbb{Q})$ directly controls the difference in expected losses via the Lipschitz constant $L$.

\paragraph{Bounding Using MMD}

Under Assumption \textbf{A4}, the function $g(x)$ lies in the RKHS $\mathcal{H}_k$ associated with the kernel $k$, with norm $\| g \|_{\mathcal{H}_k}$. The Maximum Mean Discrepancy (MMD) between $\mathbb{P}$ and $\mathbb{Q}$ is defined as \citep{gretton2012kernel}:
\begin{equation}
\mathrm{MMD}_k(\mathbb{P}, \mathbb{Q}) = \left\| \mu_{\mathbb{P}} - \mu_{\mathbb{Q}} \right\|_{\mathcal{H}_k},
\end{equation}
where $\mu_{\mathbb{P}} = \mathbb{E}_{x \sim \mathbb{P}} [ k(x, \cdot) ]$ is the mean embedding of $\mathbb{P}$ in $\mathcal{H}_k$.

Then, we have:
\begin{align}
\label{eq: mmd_bound}
\left| \mathbb{E}_{\mathbb{P}}[g(x)] - \mathbb{E}_{\mathbb{Q}}[g(x)] \right| &= \left| \langle g, \mu_{\mathbb{P}} - \mu_{\mathbb{Q}} \rangle_{\mathcal{H}_k} \right| \nonumber \\
&\leq \| g \|_{\mathcal{H}_k} \left\| \mu_{\mathbb{P}} - \mu_{\mathbb{Q}} \right\|_{\mathcal{H}_k} \nonumber \\
&= \| g \|_{\mathcal{H}_k} \, \mathrm{MMD}_k(\mathbb{P}, \mathbb{Q}).
\end{align}

\subsection{Discussion}

For a reasonably expressive neural network trained on the compact synthetic data, $\mathbb{E}_{\mathbb{Q}}[g(x)]$ should be close to $0$. From \cref{eq: wasserstein_bound} and \cref{eq: mmd_bound} we know the key in comparing the error bound for both metrics lies in comparing $L W_1(\mathbb{P}, \mathbb{Q})$ and $\| g \|_{\mathcal{H}_k} \, \mathrm{MMD}_k(\mathbb{P}, \mathbb{Q})$. When the inputs are raw pixels and $h$ includes a deep neural network, both the Lipschitz constant $L$ and the RKHS norm $\| f \|_{\mathcal{H}_k}$ can be large due to the complexity of $h$. However, when the inputs are features extracted by an encoder $e$, which is the case for most DD methods, $h$ can be a simpler function, leading to smaller values for $L$ and $\| g \|_{\mathcal{H}_k}$.

In practice, most existing MMD-based methods \cite{zhao2023dataset, wang2022cafe, zhao2023improved} approximate distribution matching by aligning only the first-order moment (mean) of the feature distributions. They minimize a loss function of the form:
\begin{equation}
\mathcal{L}_{\text{mean}} = \left\| \mu_{\mathbb{P}} - \mu_{\mathbb{Q}} \right\|^2,
\end{equation}
where $\mu_{\mathbb{P}} = \frac{1}{N} \sum_{i} g(\mathbf{x}_i)$ with $\mathbf{x}_i \sim \mathbb{P}$, and $\mu_{\mathbb{Q}} = \frac{1}{M} \sum_{j} g(\mathbf{s}_j)$ with $\mathbf{s}_j \sim \mathbb{Q}$, are the empirical means of the feature representations from the real and synthetic datasets, respectively.

This mean feature matching is mathematically equivalent to minimizing the MMD with a linear kernel:
$k(\mathbf{x}, \mathbf{y}) = \langle \mathbf{x}, \mathbf{y} \rangle$

which simplifies the MMD to:
\begin{equation}
\mathrm{MMD}_k^2(\mathbb{P}, \mathbb{Q}) = \left\| \mathbb{E}_{\mathbf{x} \sim \mathbb{P}}[\mathbf{x}] - \mathbb{E}_{\mathbf{x} \sim \mathbb{Q}}[\mathbf{x}] \right\|^2.
\end{equation}

However, the linear kernel is generally not characteristic, meaning it cannot uniquely distinguish all probability distributions. As a result, aligning only the means leads to inaccurate distribution matching, neglecting higher-order moments like variance and skewness. This inaccuracy can cause the actual discrepancy between the distributions to remain large, even if the MMD computed with the linear kernel is minimized.
Consequently, the inaccurate approximation does not reduce the actual MMD value that would be computed with a characteristic kernel, leaving a significant distributional mismatch unaddressed.

The M3D method \cite{zhang2024m3d} improves the precision of MMD-based distribution matching by using a more expressive kernel such as the Gaussian RBF kernel, which effectively captures discrepancies across all moments, with the MMD equation below:
\begin{align}
\mathrm{MMD}_k^2(\mathbb{P}, \mathbb{Q}) =& \mathbb{E}_{\mathbf{x}, \mathbf{x}' \sim \mathbb{P}}[k(\mathbf{x}, \mathbf{x}')] + \mathbb{E}_{\mathbf{s}, \mathbf{s}' \sim \mathbb{Q}}[k(\mathbf{s}, \mathbf{s}')] \\
&- 2\mathbb{E}_{\mathbf{x} \sim \mathbb{P}, \mathbf{s} \sim \mathbb{Q}}[k(\mathbf{x}, \mathbf{s})].
\end{align}

However, this approach introduces sensitivity to the choice of kernel and its parameters, which may be less favorable because an unsuitable kernel may fail to capture important characteristics of the distributions. Moreover, computing the full MMD with a characteristic kernel requires evaluating the kernel function for all pairs of data points, including those from the extensive real dataset, scaling quadratically with dataset size. As a result, this method generally incurs more computational cost compared to earlier methods such as DM and does not scale to large datasets such as ImageNet-1K.

In general, existing MMD-based methods often struggle to achieve precise distribution matching in a way scalable to large datasets. In contrast, the Wasserstein-1 distance inherently accounts for discrepancies in all moments without relying on a kernel function. Its computational feasibility is ensured by the efficient algorithms for Wasserstein barycenter computation and the reduced dimensionality in the feature space. This may explain why, in our experiments, the Wasserstein-1 distance led to better performance than MMD-based approaches that rely on mean feature matching with linear kernels.

\section{More Explanations on the Method}
\label{sec: math_detail}
In this section, we expand our discussion in \cref{sec: bary_compute} in more details, to explain how we adapt the method in \citep{cuturi2014fast} for efficient computation of the Wasserstein barycenter.
\subsection{Optimizing Weights Given Fixed Positions}
The optimization of weights given fixed positions in the optimal transport problem involves solving a linear programming (LP) problem, where the primal form seeks the minimal total transportation cost subject to constraints on mass distribution. Given the cost matrix \(C\) and the transport plan \(T\), the primal problem is formulated as:
\begin{gather}
\min_{\mathbf{T}} \langle C, \mathbf{T} \rangle_F \quad  \label{eq: cost2} \\
\text{subject to} \quad \sum_{j=1}^{m} t_{ij} = \frac{1}{n}, \; \forall i, \quad \\
\sum_{i=1}^{n} t_{ij} = w_j,  \; \forall j, \quad  t_{ij} \geq 0, \; \forall i, j,
\end{gather}
where \(\langle \cdot, \cdot \rangle_F\) denotes the Frobenius inner product.

The corresponding dual problem introduces dual variables \(\alpha_i\) and \(\beta_j\), maximizing the objective:
\begin{gather}
\max_{\alpha, \beta} \left\{ \sum_{i=1}^{n} \frac{\alpha_i}{n} + \sum_{j=1}^{m} w_j \beta_j \right\}\\
\textrm{subject to} \quad \alpha_i + \beta_j \leq c_{ij}, \forall i, j.
\end{gather}

Given the LP's feasibility and boundedness, strong duality holds, confirming that both the primal and dual problems reach the same optimal value \citep{boyd2004convex}. This equivalence implies that the set of optimal dual variables denoted as $\boldsymbol{\beta}$, acts as a subgradient, guiding the weight updates. Specifically, this subgradient indicates how the marginal costs vary with changes in the weights. To update the weights $\mathbf{w}$ towards their optimal values $\mathbf{w}^\star$, we implement the projected subgradient descent technique. This method ensures that $\mathbf{w}$ remains within the probability simplex, and under appropriate conditions on the step sizes, it guarantees convergence to the optimal solution.
\subsection{Optimizing Positions Given Fixed Weights}

\subsubsection{Gradient Computation}
Given the cost matrix \(C\) with elements \(c_{ij} = \|\tilde{\mathbf{x}}_j - \mathbf{x}_i\|^2\), the gradient of the cost function with respect to a synthetic position \(\tilde{\mathbf{x}}_j\) is derived from the partial derivatives of \(c_{ij}\) with respect to \(\tilde{\mathbf{x}}_j\). The gradient of \(c_{ij}\) with respect to \(\tilde{\mathbf{x}}_j\) is:
\begin{align}
\nabla_{\tilde{\mathbf{x}}_j} c_{ij} = 2(\tilde{\mathbf{x}}_j - \mathbf{x}_i).
\end{align}

However, the overall gradient depends on the transport plan \(\mathbf{T}\) that solves the optimal transport problem. The gradient of the cost function \(f\) with respect to \(\tilde{\mathbf{x}}_j\) takes into account the amount of mass \(t_{ij}\) transported from \(\tilde{\mathbf{x}}_j\) to \(\mathbf{x}_i\):
\begin{align}
\nabla_{\tilde{\mathbf{x}}_j} f(\tilde{\mathbf{X}}) = \sum_{i=1}^{n} t_{ij} \nabla_{\tilde{\mathbf{x}}_j} c_{ij} = \sum_{i=1}^{n} t_{ij} 2(\tilde{\mathbf{x}}_j - \mathbf{x}_i).
\end{align}

\begin{algorithm}[h]
\SetAlgoLined
\KwResult{Optimized barycenter matrix $\mathbf{B}^*$ and weights $\mathbf{w}^*$.}
\textbf{Input:} Feature matrix of real data $\mathbf{Z} \in \mathbb{R}^{n_k \times d_f}$, initial synthetic dataset positions $\mathbf{B}^{(0)} \in \mathbb{R}^{m_k \times d_f}$, number of iterations $K$, learning rate $\eta$\;
Initialize weights $\mathbf{w}^{(0)}$ uniformly\;
\For{$k = 1$ \KwTo $K$}{
    \tcp{Optimize weights given positions}
    Construct cost matrix $C^{(k)}$ with $\mathbf{B}^{(k-1)}$ and $\mathbf{Z}$\;
    Solve optimal transport problem to obtain transport plan $\mathbf{T}^{(k)}$ and dual variables $\boldsymbol{\beta}^{(k)}$\;
    Update weights $\mathbf{w}^{(k)}$ using projected subgradient method: $\mathbf{w}^{(k)} = \text{Project}\left(\mathbf{w}^{(k-1)} - \eta \boldsymbol{\beta}^{(k)}\right)$, ensuring $w_j^{(k)} \geq 0$ and $\sum_j w_j^{(k)} = 1$\;
    
    \tcp{Optimize positions given weights}
    Compute gradient $\nabla_{\mathbf{B}} f$ as per: $\nabla_{\mathbf{b}_{j}} f = \sum_{i=1}^{n} t_{ij}^{(k)} 2(\mathbf{b}_{j}^{(k-1)} - \mathbf{z}_i), \forall j$\;
    Update positions $\mathbf{B}^{(k)}$ using Newton's method: $\mathbf{b}_{j}^{(k)} = \mathbf{b}_{j}^{(k-1)} - H_j^{-1} \nabla_{\mathbf{b}_{j}} f, \forall j$, where $H_j$ is the Hessian\;
}
$\mathbf{B}^* \leftarrow \mathbf{B}^{(K)}$, $\mathbf{w}^* \leftarrow \mathbf{w}^{(K)}$\;
\caption{Iterative Barycenter Learning for Dataset Distillation\label{alg: bary_learning}}
\end{algorithm}

\subsubsection{Hessian Computation}

The Hessian matrix \(H\) of \(f\) with respect to \(\tilde{\mathbf{X}}\) involves second-order partial derivatives. For \(p=2\), the second-order partial derivative of \(c_{ij}\) with respect to \(\tilde{\mathbf{x}}_j\) is constant:
\begin{align}
\frac{\partial^2 c_{ij}}{\partial \tilde{\mathbf{x}}_j^2} = 2\mathbf{I},
\end{align}
where \(\mathbf{I}\) is the identity matrix. Thus, the Hessian of \(f\) with respect to \(\tilde{\mathbf{X}}\) for each synthetic point \(\tilde{\mathbf{x}}_j\) is:
\begin{align}
H_j = \sum_{i=1}^{n} t_{ij} 2\mathbf{I} = 2\mathbf{I} \sum_{i=1}^{n} t_{ij} = 2\mathbf{I}w_j,
\end{align}
since \(\sum_{i=1}^{n} t_{ij} = w_j\), the amount of mass associated with synthetic point \(\tilde{\mathbf{x}}_j\).

\subsubsection{Newton Update Formula}

The Newton update formula for each synthetic position \(\tilde{\mathbf{x}}_j\) is then:
\begin{align}
\tilde{\mathbf{x}}_j^{(\text{new})} &= \tilde{\mathbf{x}}_j - H_j^{-1} \nabla_{\tilde{\mathbf{x}}_j} f(\tilde{\mathbf{X}}) \\
&= \tilde{\mathbf{x}}_j - \frac{1}{2w_j} \sum_{i=1}^{n} t_{ij} 2(\tilde{\mathbf{x}}_j - \mathbf{x}_i).
\end{align}

Simplifying, we obtain:
\begin{align}
\tilde{\mathbf{x}}_j^{(\text{new})} = \tilde{\mathbf{x}}_j - \sum_{i=1}^{n} t_{ij} (\tilde{\mathbf{x}}_j - \mathbf{x}_i) / w_j.
\end{align}

This formula adjusts each synthetic position \(\tilde{\mathbf{x}}_j\) in the direction that reduces the Wasserstein distance, weighted by the amount of mass transported and normalized by the weight \(w_j\).

\section{Algorithm details}
\label{sec: alg_detail}
As discussed in \cref{sec: engi} (Algorithm \ref{alg: wasserstein_barycenter}) in the main paper, our method involves 
computing the Wasserstein barycenter of the empirical distribution of intra-class features. This section details the algorithm 
employed.

Let us denote the training set as $\mathcal{T} = \{\mathbf{x}_{k, i}\}_{i=1, \ldots, n_k}^{k=1, \ldots, g}$, where $g$ is the 
number of classes and $n_k$ is the number of images in class $k$. In the rest of this section, we only discuss the computation 
for class $k$, so we omit the index $k$ from the subscript of related symbols for simplicity, e.g., $\mathbf{x}_{k, i}$ is 
simplified as $\mathbf{x}_{i}$. A feature extractor $f_e(\cdot)$ embeds the real data of this class into the feature space 
$\mathbb{R}^{d_f}$, yielding a feature matrix $\mathbf{Z} \in \mathbb{R}^{n_k \times d_f}$, where the $i$th row $\mathbf{z}_i = f_e(\mathbf{x}_{k, i})$. 
We employ the algorithm shown in Algorithm \ref{alg: bary_learning} to compute the Wasserstein barycenter of the feature distribution. 
It takes $\mathbf{Z}$ as input and outputs a barycenter matrix $\mathbf{B}^* \in \mathbb{R}^{m_k \times d_f}$, where the $j$th row $\mathbf{b}_{j}^*$ 
is the feature for learning the $j$th synthetic image, and an associated weight vector (probability distribution) $\mathbf{w}^*\in \mathbb{R}^{m_k}$. 
\vspace{5mm}

\section{Implementation details}
\label{sec: impl_detail}
In our experiments, each experiment run was conducted on a single GPU of type A40, A100, or RTX-3090, depending on the availability. We used torchvision \citep{torchvision2016} for pretraining of models in the squeeze stage, and slightly modified the model architecture to allow tracking of per-class BatchNorm statistics. 

We remained most of the hyperparameters in~\citep{yin2023squeeze} despite a few modifications. In the squeeze stage, we reduced the batch size to $32$ for single-GPU training and correspondingly reduced the learning rate to $0.025$. In addition, we find from preliminary experiments that the weight decay at the recovery stage is detrimental to the performance of synthetic data, so we set them to $0$.  

For our loss term in \cref{eq: total_loss}, we set lambda ($\lambda$) to $500$ for ImageNet, $300$ for Tiny-ImageNet, and $10$ for ImageNette. We set the number of iterations to $2000$ for all datasets. Table~\ref{tab:hyp_imagenette}-\ref{tab:hyp_1k} shows the hyperparameters used in the recover stage of our method. Hyperparameters in subsequent stages are kept the same as in~\citep{yin2023squeeze}.

\begin{table*}[htbp!]
\centering
\small
\setlength{\tabcolsep}{1mm}
\begin{minipage}[t]{0.24\textwidth}
\centering
\begin{tabular}{@{}lc@{}}
\toprule
\multicolumn{1}{c}{\textbf{config}} & \textbf{value} \\ \midrule
optimizer           & SGD            \\
learning rate  & 0.025            \\
weight decay        & 1e-4           \\
opti. mom.  & 0.9            \\
batch size          & 32            \\
scheduler & cosine decay \\
train. epoch      & 100            \\ \bottomrule
\end{tabular}
\subcaption{Squeezing setting for all datasets}
\end{minipage}
\hspace{0pt}
\begin{minipage}[t]{0.24\textwidth}
\centering
\begin{tabular}{@{}lc@{}}
\toprule
\multicolumn{1}{c}{\textbf{config}} & \textbf{value} \\ \midrule
lambda    & 10            \\
optimizer          & Adam           \\
learning rate & 0.25            \\
opti. mom. & \(\beta_1, \beta_2 = 0.5, 0.9\) \\
batch size         & 100            \\
scheduler & cosine decay \\
recover. iter. & 2,000        \\ \bottomrule
\end{tabular}
\subcaption{Recovering setting for ImageNette}
\label{tab:hyp_imagenette}
\end{minipage}
\hspace{0pt}
\begin{minipage}[t]{0.24\textwidth}
\centering
\begin{tabular}{@{}lc@{}}
\toprule
\multicolumn{1}{c}{\textbf{config}} & \textbf{value} \\ \midrule
lambda    & 300            \\
optimizer          & Adam           \\
learning rate & 0.1            \\
opti. mom. & \(\beta_1, \beta_2 = 0.5, 0.9\) \\
batch size         & 100            \\
scheduler & cosine decay \\
recover. iter. & 2,000        \\ \bottomrule
\end{tabular}
\subcaption{Recovering setting for Tiny-ImageNet}
\label{tab:hyp_tiny}
\end{minipage}
\hspace{0pt}
\begin{minipage}[t]{0.24\textwidth}
\centering
\begin{tabular}{@{}lc@{}}
\toprule
\multicolumn{1}{c}{\textbf{config}} & \textbf{value} \\ \midrule
lambda    & 500            \\
optimizer          & Adam           \\
learning rate & 0.25            \\
opti. mom. & \(\beta_1, \beta_2 = 0.5, 0.9\) \\
batch size         & 100            \\
scheduler & cosine decay \\
recover. iter. & 2,000        \\
\bottomrule
\end{tabular}
\subcaption{Recovering setting for ImageNet-1K}
\label{tab:hyp_1k}
\end{minipage}
\caption{Hyperparameter settings for model training and recovering.}
\end{table*}

\begin{figure*}[hb]
    \centering
        \hfill
        \hfill
        \hfill
        \hfill
        \hfill
        \hfill
        \hfill

    \begin{adjustbox}{width=\textwidth,center}
        \begin{subfigure}[b]{0.1\linewidth}
            \centering
            \includegraphics[width=\linewidth]{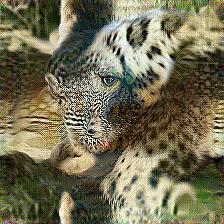}
            \caption*{\scriptsize Leopard}
        \end{subfigure}
        \hfill
        \begin{subfigure}[b]{0.1\linewidth}
            \centering
            \includegraphics[width=\linewidth]{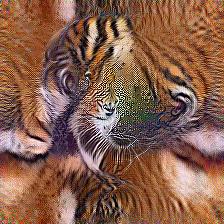}
            \caption*{\scriptsize Tiger}
        \end{subfigure}
        \hfill
        \begin{subfigure}[b]{0.1\linewidth}
            \centering
            \includegraphics[width=\linewidth]{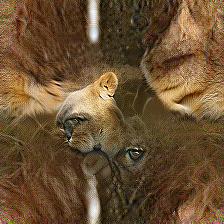}
            \caption*{\scriptsize Lion}
        \end{subfigure}
        \hfill
        \begin{subfigure}[b]{0.1\linewidth}
            \centering
            \includegraphics[width=\linewidth]{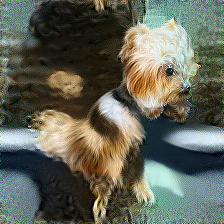}
            \caption*{\scriptsize Yorkshire}
        \end{subfigure}
        \hfill
        \begin{subfigure}[b]{0.1\linewidth}
            \centering
            \includegraphics[width=\linewidth]{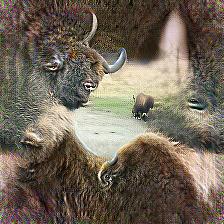}
            \caption*{\scriptsize Bison}
        \end{subfigure}
        \hfill
        \begin{subfigure}[b]{0.1\linewidth}
            \centering
            \includegraphics[width=\linewidth]{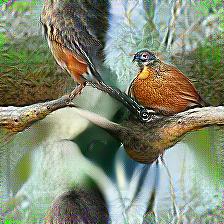}
            \caption*{\scriptsize Robin}
        \end{subfigure}
        \hfill
        \begin{subfigure}[b]{0.1\linewidth}
            \centering
            \includegraphics[width=\linewidth]{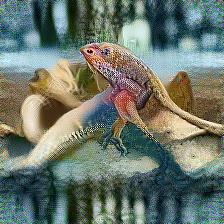}
            \caption*{\scriptsize Agama}
        \end{subfigure}
        \hfill
        \begin{subfigure}[b]{0.1\linewidth}
            \centering
            \includegraphics[width=\linewidth]{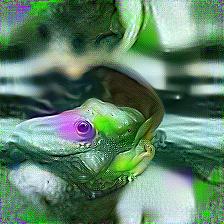}
            \caption*{\scriptsize Tree Frog}
        \end{subfigure}
        \hfill
        \begin{subfigure}[b]{0.1\linewidth}
            \centering
            \includegraphics[width=\linewidth]{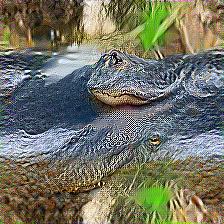}
            \caption*{\scriptsize Alligator}
        \end{subfigure}
        \hfill
        \begin{subfigure}[b]{0.1\linewidth}
            \centering
            \includegraphics[width=\linewidth]{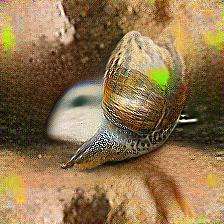}
            \caption*{\scriptsize Snail}
        \end{subfigure}
    \end{adjustbox}

    \begin{adjustbox}{width=\textwidth,center}
        \begin{subfigure}[b]{0.1\linewidth}
            \centering
            \includegraphics[width=\linewidth]{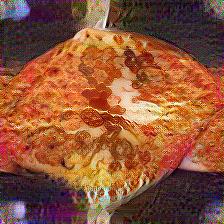}
            \caption*{\scriptsize Pizza}
        \end{subfigure}
        \hfill
        \begin{subfigure}[b]{0.1\linewidth}
            \centering
            \includegraphics[width=\linewidth]{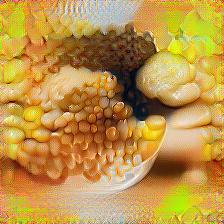}
            \caption*{\scriptsize Corn}
        \end{subfigure}
        \hfill
        \begin{subfigure}[b]{0.1\linewidth}
            \centering
            \includegraphics[width=\linewidth]{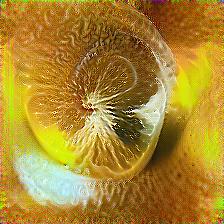}
            \caption*{\scriptsize Lemon}
        \end{subfigure}
        \hfill
        \begin{subfigure}[b]{0.1\linewidth}
            \centering
            \includegraphics[width=\linewidth]{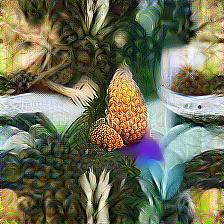}
            \caption*{\scriptsize Pineapple}
        \end{subfigure}
        \hfill
        \begin{subfigure}[b]{0.1\linewidth}
            \centering
            \includegraphics[width=\linewidth]{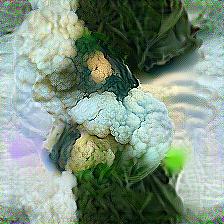}
            \caption*{\scriptsize Cauliflower}
        \end{subfigure}
        \hfill
        \begin{subfigure}[b]{0.1\linewidth}
            \centering
            \includegraphics[width=\linewidth]{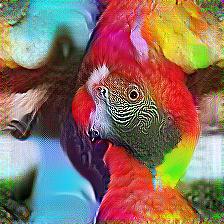}
            \caption*{\scriptsize Macaw}
        \end{subfigure}
        \hfill
        \begin{subfigure}[b]{0.1\linewidth}
            \centering
            \includegraphics[width=\linewidth]{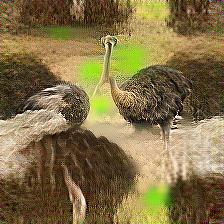}
            \caption*{\scriptsize Ostrich}
        \end{subfigure}
        \hfill
        \begin{subfigure}[b]{0.1\linewidth}
            \centering
            \includegraphics[width=\linewidth]{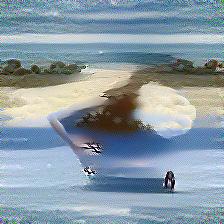}
            \caption*{\scriptsize Seashore}
        \end{subfigure}
        \hfill
        \begin{subfigure}[b]{0.1\linewidth}
            \centering
            \includegraphics[width=\linewidth]{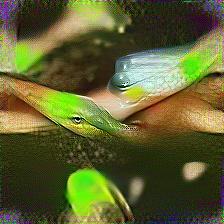}
            \caption*{\scriptsize Snake}
        \end{subfigure}
        \hfill
        \begin{subfigure}[b]{0.1\linewidth}
            \centering
            \includegraphics[width=\linewidth]{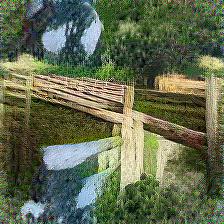}
            \caption*{\scriptsize Fence}
        \end{subfigure}
        
    \end{adjustbox}
    
    \begin{adjustbox}{width=\textwidth,center}
        \setlength{\fboxsep}{0pt}
        \begin{subfigure}[b]{0.1\linewidth}
            \centering
            \includegraphics[width=\linewidth]{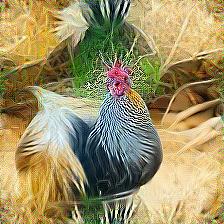}
            \caption*{\scriptsize Cock}
        \end{subfigure}
        \hfill
        \begin{subfigure}[b]{0.1\linewidth}
            \centering
            \includegraphics[width=\linewidth]{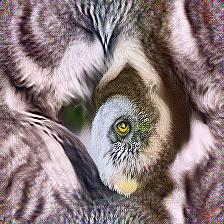}
            \caption*{\scriptsize Grey Owl}
        \end{subfigure}
        \hfill
        \begin{subfigure}[b]{0.1\linewidth}
            \centering
            \includegraphics[width=\linewidth]{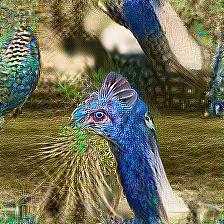}
            \caption*{\scriptsize Peacock}
        \end{subfigure}
        \hfill
        \begin{subfigure}[b]{0.1\linewidth}
            \centering
            \includegraphics[width=\linewidth]{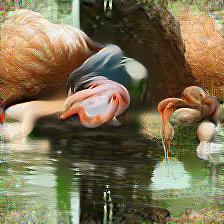}
            \caption*{\scriptsize Flamingo}
        \end{subfigure}
        \hfill
        \begin{subfigure}[b]{0.1\linewidth}
            \centering
            \includegraphics[width=\linewidth]{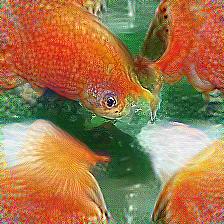}
            \caption*{\scriptsize Gold Fish}
        \end{subfigure}
        \begin{subfigure}[b]{0.1\linewidth}
            \centering
            \includegraphics[width=\linewidth]{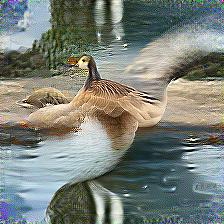}
            \caption*{\scriptsize Goose}
        \end{subfigure}
        \hfill
        \begin{subfigure}[b]{0.1\linewidth}
            \centering
            \includegraphics[width=\linewidth]{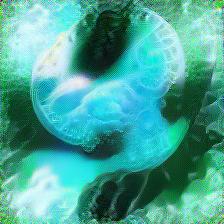}
            \caption*{\scriptsize Jellyfish}
        \end{subfigure}
        \hfill
        \begin{subfigure}[b]{0.1\linewidth}
            \centering
            \includegraphics[width=\linewidth]{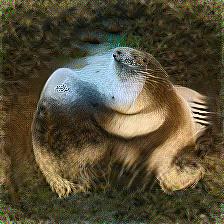}
            \caption*{\scriptsize Sea Lion}
        \end{subfigure}
        \hfill
        \begin{subfigure}[b]{0.1\linewidth}
            \centering
            \includegraphics[width=\linewidth]{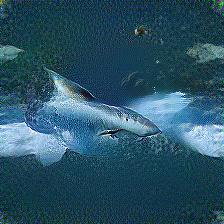}
            \caption*{\scriptsize Shark}
        \end{subfigure}
        \hfill
        \begin{subfigure}[b]{0.1\linewidth}
            \centering
            \includegraphics[width=\linewidth]{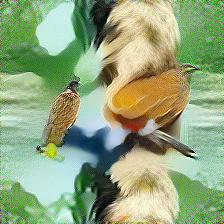}
            \caption*{\scriptsize Bulbul}
        \end{subfigure}
    \end{adjustbox}
    \begin{adjustbox}{width=\textwidth,center}
        \setlength{\fboxsep}{0pt}
        \begin{subfigure}[b]{0.1\linewidth}
            \centering
            \includegraphics[width=\linewidth]{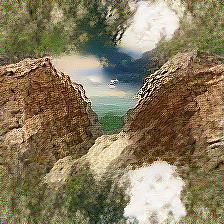}
            \caption*{\scriptsize Cliff}
        \end{subfigure}
        \hfill
        \begin{subfigure}[b]{0.1\linewidth}
            \centering
            \includegraphics[width=\linewidth]{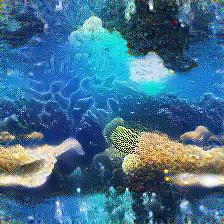}
            \caption*{\scriptsize CoralReef}
        \end{subfigure}
        \hfill
        \begin{subfigure}[b]{0.1\linewidth}
            \centering
            \includegraphics[width=\linewidth]{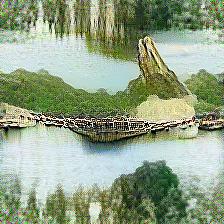}
            \caption*{\scriptsize Lakeside}
        \end{subfigure}
        \hfill
        \begin{subfigure}[b]{0.1\linewidth}
            \centering
            \includegraphics[width=\linewidth]{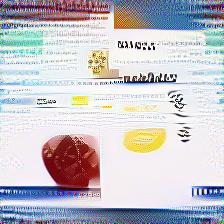}
            \caption*{\scriptsize Website}
        \end{subfigure}
        \hfill
        \begin{subfigure}[b]{0.1\linewidth}
            \centering
            \includegraphics[width=\linewidth]{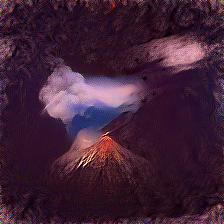}
            \caption*{\scriptsize Volcano}
        \end{subfigure}
        \begin{subfigure}[b]{0.1\linewidth}
            \centering
            \includegraphics[width=\linewidth]{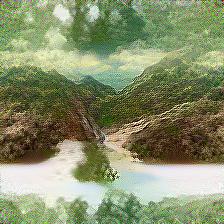}
            \caption*{\scriptsize Valley}
        \end{subfigure}
        \hfill
        \begin{subfigure}[b]{0.1\linewidth}
            \centering
            \includegraphics[width=\linewidth]{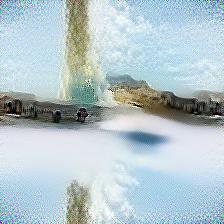}
            \caption*{\scriptsize Geyser}
        \end{subfigure}
        \hfill
        \begin{subfigure}[b]{0.1\linewidth}
            \centering
            \includegraphics[width=\linewidth]{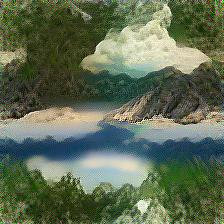}
            \caption*{\scriptsize Foreland}
        \end{subfigure}
        \hfill
        \begin{subfigure}[b]{0.1\linewidth}
            \centering
            \includegraphics[width=\linewidth]{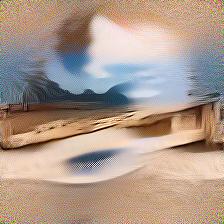}
            \caption*{\scriptsize Sandbar}
        \end{subfigure}
        \hfill
        \begin{subfigure}[b]{0.1\linewidth}
            \centering
            \includegraphics[width=\linewidth]{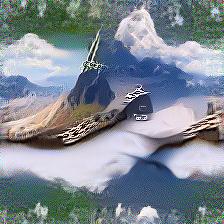}
            \caption*{\scriptsize Alp}
        \end{subfigure}
        \end{adjustbox}
        \begin{adjustbox}{width=\linewidth,center}
        \setlength{\fboxsep}{0pt}
        \begin{subfigure}[b]{0.1\linewidth}
            \centering
            \includegraphics[width=\linewidth]{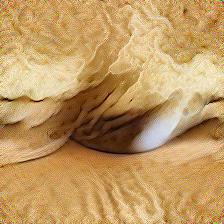}
            \caption*{\scriptsize Dough}
        \end{subfigure}
        \hfill
        \begin{subfigure}[b]{0.1\linewidth}
            \centering
            \includegraphics[width=\linewidth]{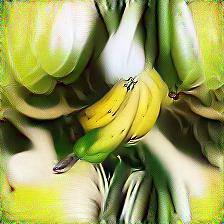}
            \caption*{\scriptsize Banana}
        \end{subfigure}
        \hfill
        \begin{subfigure}[b]{0.1\linewidth}
            \centering
            \includegraphics[width=\linewidth]{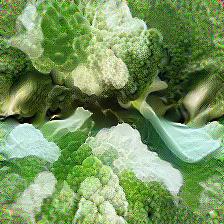}
            \caption*{\scriptsize Broccoli}
        \end{subfigure}
        \hfill
        \begin{subfigure}[b]{0.1\linewidth}
            \centering
            \includegraphics[width=\linewidth]{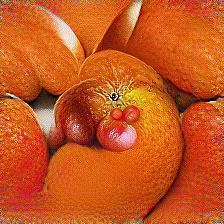}
            \caption*{\scriptsize Orange}
        \end{subfigure}
        \hfill
        \begin{subfigure}[b]{0.1\linewidth}
            \centering
            \includegraphics[width=\linewidth]{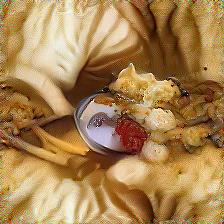}
            \caption*{\scriptsize Potato}
        \end{subfigure}
        \hfill
        \begin{subfigure}[b]{0.1\linewidth}
            \centering
            \includegraphics[width=\linewidth]{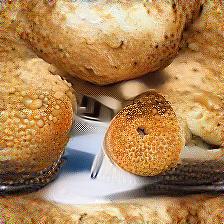}
            \caption*{\scriptsize Bagel}
        \end{subfigure}
        \hfill
        \begin{subfigure}[b]{0.1\linewidth}
            \centering
            \includegraphics[width=\linewidth]{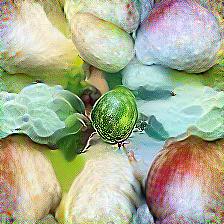}
            \caption*{\scriptsize Fig}
        \end{subfigure}
        \hfill
        \begin{subfigure}[b]{0.1\linewidth}
            \centering
            \includegraphics[width=\linewidth]{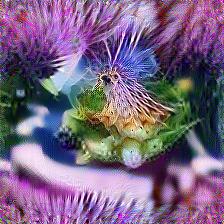}
            \caption*{\scriptsize Cardoon}
        \end{subfigure}
        \hfill
        \begin{subfigure}[b]{0.1\linewidth}
            \centering
            \includegraphics[width=\linewidth]{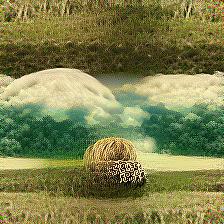}
            \caption*{\scriptsize Hay}
        \end{subfigure}
        \hfill
        \begin{subfigure}[b]{0.1\linewidth}
            \centering
            \includegraphics[width=\linewidth]{figs/Imagenet_our_v2_10ipc/class966_id008.jpg}
            \caption*{\scriptsize Red Wine}
        \end{subfigure}
    \end{adjustbox}
    \caption{Visualizations of our synthetic images from ImageNet-1K}
    \label{fig:visualization-1K}
\end{figure*}

\begin{figure*}[tb]
    \centering
    \begin{adjustbox}{width=\linewidth,center}
        \begin{subfigure}[b]{0.1\linewidth}
            \centering
            \includegraphics[width=\linewidth]{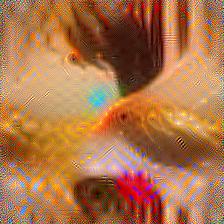}
            \caption*{\scriptsize Goldfish}
        \end{subfigure}
        \hfill
        \begin{subfigure}[b]{0.1\linewidth}
            \centering
            \includegraphics[width=\linewidth]{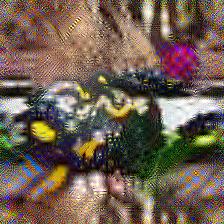}
            \caption*{\scriptsize Salamander}
        \end{subfigure}
        \hfill
        \begin{subfigure}[b]{0.1\linewidth}
            \centering
            \includegraphics[width=\linewidth]{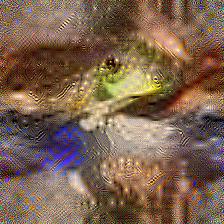}
            \caption*{\scriptsize Bullfrog}
        \end{subfigure}
        \hfill
        \begin{subfigure}[b]{0.1\linewidth}
            \centering
            \includegraphics[width=\linewidth]{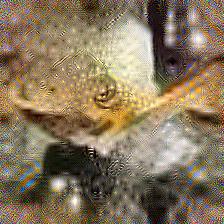}
            \caption*{\scriptsize TailedFrog}
        \end{subfigure}
        \hfill
        \begin{subfigure}[b]{0.1\linewidth}
            \centering
            \includegraphics[width=\linewidth]{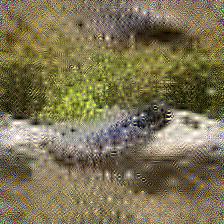}
            \caption*{\scriptsize Alligator}
        \end{subfigure}
        \hfill
        \begin{subfigure}[b]{0.1\linewidth}
            \centering
            \includegraphics[width=\linewidth]{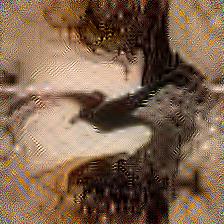}
            \caption*{\scriptsize Scorpion}
        \end{subfigure}
        \hfill
        \begin{subfigure}[b]{0.1\linewidth}
            \centering
            \includegraphics[width=\linewidth]{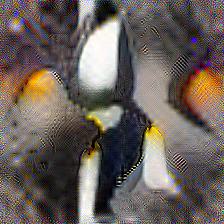}
            \caption*{\scriptsize Penguin}
        \end{subfigure}
        \hfill
        \begin{subfigure}[b]{0.1\linewidth}
            \centering
            \includegraphics[width=\linewidth]{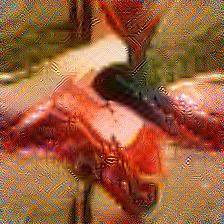}
            \caption*{\scriptsize Lobster}
        \end{subfigure}
        \hfill
        \begin{subfigure}[b]{0.1\linewidth}
            \centering
            \includegraphics[width=\linewidth]{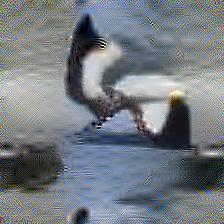}
            \caption*{\scriptsize Sea Gull}
        \end{subfigure}
        \hfill
        \begin{subfigure}[b]{0.1\linewidth}
            \centering
            \includegraphics[width=\linewidth]{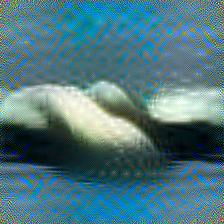}
            \caption*{\scriptsize Sea Lion}
        \end{subfigure}
    \end{adjustbox}
    \caption*{Tiny-ImageNet}

    \begin{adjustbox}{width=\linewidth,center}
        \begin{subfigure}[b]{0.1\linewidth}
            \centering
            \includegraphics[width=\linewidth]{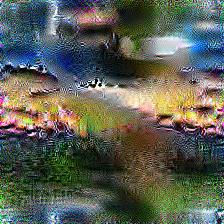}
            \caption*{\scriptsize Tench}
        \end{subfigure}
        \hfill
        \begin{subfigure}[b]{0.1\linewidth}
            \centering
            \includegraphics[width=\linewidth]{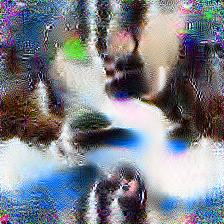}
            \caption*{\scriptsize Springer}
        \end{subfigure}
        \hfill
        \begin{subfigure}[b]{0.1\linewidth}
            \centering
            \includegraphics[width=\linewidth]{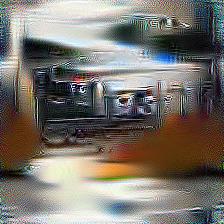}
            \caption*{\scriptsize CassettePlyr}
        \end{subfigure}
        \hfill
        \begin{subfigure}[b]{0.1\linewidth}
            \centering
            \includegraphics[width=\linewidth]{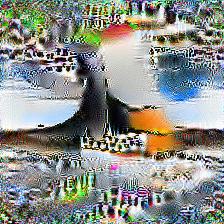}
            \caption*{\scriptsize Chain Saw}
        \end{subfigure}
        \hfill
        \begin{subfigure}[b]{0.1\linewidth}
            \centering
            \includegraphics[width=\linewidth]{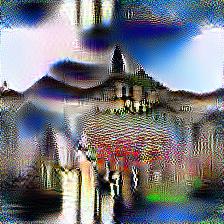}
            \caption*{\scriptsize Church}
        \end{subfigure}
        \hfill
        \begin{subfigure}[b]{0.1\linewidth}
            \centering
            \includegraphics[width=\linewidth]{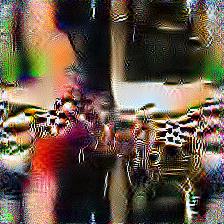}
            \caption*{\scriptsize FrenchHorn}
        \end{subfigure}
        \hfill
        \begin{subfigure}[b]{0.1\linewidth}
            \centering
            \includegraphics[width=\linewidth]{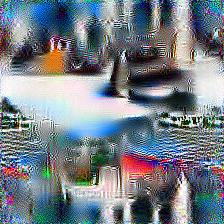}
            \caption*{\scriptsize Garb. Truck}
        \end{subfigure}
        \hfill
        \begin{subfigure}[b]{0.1\linewidth}
            \centering
            \includegraphics[width=\linewidth]{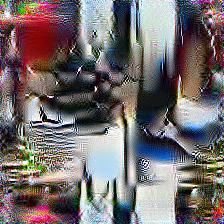}
            \caption*{\scriptsize Gas Pump}
        \end{subfigure}
        \hfill
        \begin{subfigure}[b]{0.1\linewidth}
            \centering
            \includegraphics[width=\linewidth]{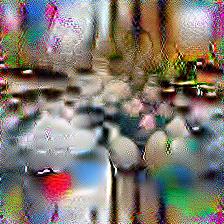}
            \caption*{\scriptsize Golf Ball}
        \end{subfigure}
        \hfill
        \begin{subfigure}[b]{0.1\linewidth}
            \centering
            \includegraphics[width=\linewidth]{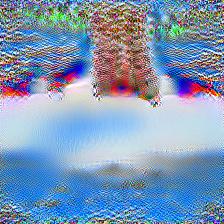}
            \caption*{\scriptsize Parachute}
        \end{subfigure}
    \end{adjustbox}
    \caption*{ImageNette}

    \caption{Visualizations of our synthetic images on smaller datasets}
    \label{fig:Synthetic_Image_smaller}
\end{figure*}

\begin{figure*}
    \centering
    \begin{adjustbox}{width=\linewidth,center}
    \renewcommand{\arraystretch}{1}
    \setlength{\fboxsep}{0pt}
    \begin{tabular}{@{}*{10}{m{0.1\textwidth}@{}}}
        \fbox{\includegraphics[width=0.95\linewidth]{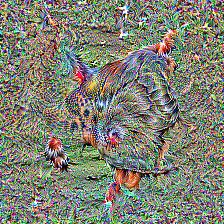}} &
        \fbox{\includegraphics[width=0.95\linewidth]{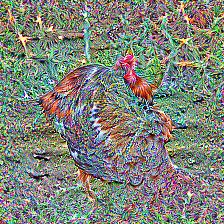}} &
        \fbox{\includegraphics[width=0.95\linewidth]{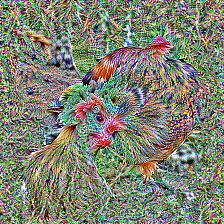}} &
        \fbox{\includegraphics[width=0.95\linewidth]{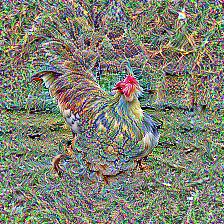}} &
        \fbox{\includegraphics[width=0.95\linewidth]{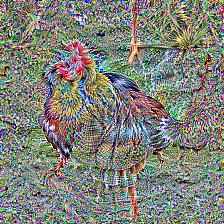}} &
        \fbox{\includegraphics[width=0.95\linewidth]{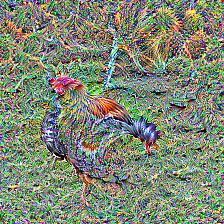}} &
        \fbox{\includegraphics[width=0.95\linewidth]{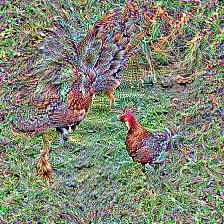}} &
        \fbox{\includegraphics[width=0.95\linewidth]{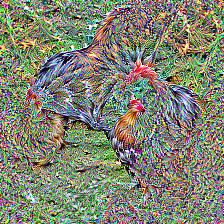}} &
        \fbox{\includegraphics[width=0.95\linewidth]{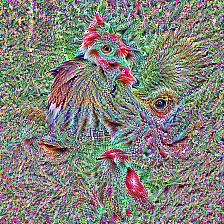}} &
        \fbox{\includegraphics[width=0.95\linewidth]{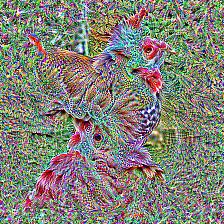}}
        \\
        \multicolumn{10}{c}{\footnotesize $\lambda=0.1$} \\
        \fbox{\includegraphics[width=0.95\linewidth]{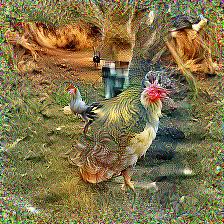}} &
        \fbox{\includegraphics[width=0.95\linewidth]{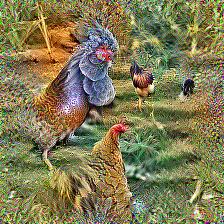}} &
        \fbox{\includegraphics[width=0.95\linewidth]{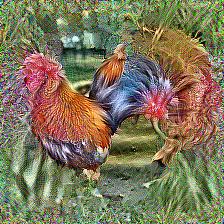}} &
        \fbox{\includegraphics[width=0.95\linewidth]{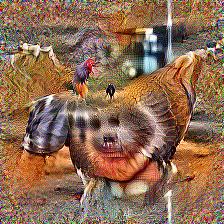}} &
        \fbox{\includegraphics[width=0.95\linewidth]{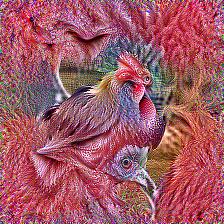}} &
        \fbox{\includegraphics[width=0.95\linewidth]{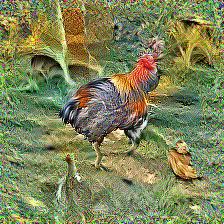}} &
        \fbox{\includegraphics[width=0.95\linewidth]{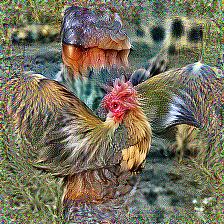}} &
        \fbox{\includegraphics[width=0.95\linewidth]{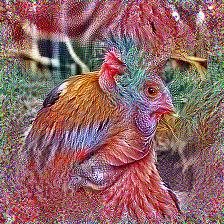}} &
        \fbox{\includegraphics[width=0.95\linewidth]{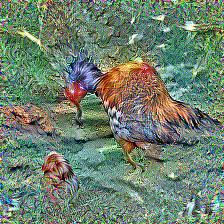}} &
        \fbox{\includegraphics[width=0.95\linewidth]{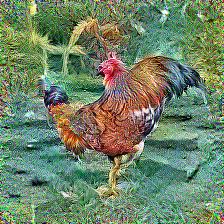}}
        \\
        \multicolumn{10}{c}{\footnotesize $\lambda=1$} \\
        \fbox{\includegraphics[width=0.95\linewidth]{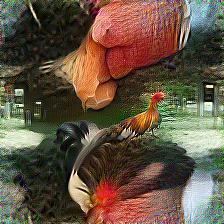}} &
        \fbox{\includegraphics[width=0.95\linewidth]{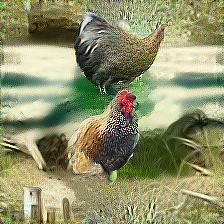}} &
        \fbox{\includegraphics[width=0.95\linewidth]{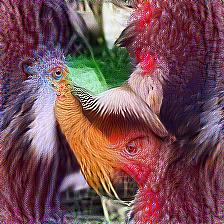}} &
        \fbox{\includegraphics[width=0.95\linewidth]{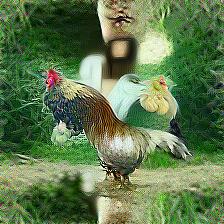}} &
        \fbox{\includegraphics[width=0.95\linewidth]{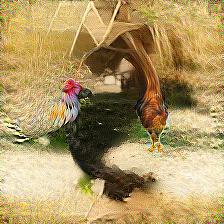}} &
        \fbox{\includegraphics[width=0.95\linewidth]{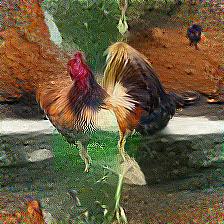}} &
        \fbox{\includegraphics[width=0.95\linewidth]{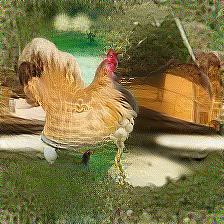}} &
        \fbox{\includegraphics[width=0.95\linewidth]{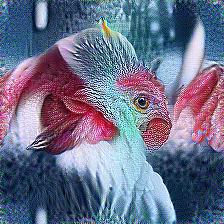}} &
        \fbox{\includegraphics[width=0.95\linewidth]{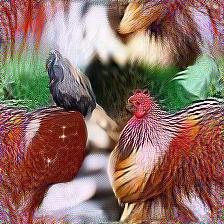}} &
        \fbox{\includegraphics[width=0.95\linewidth]{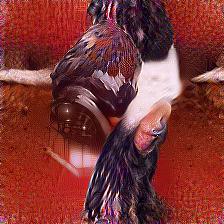}}
        \\
        \multicolumn{10}{c}{\footnotesize $\lambda=10$} \\
        \fbox{\includegraphics[width=0.95\linewidth]{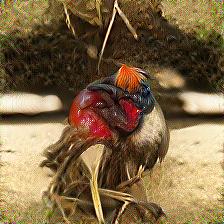}} &
        \fbox{\includegraphics[width=0.95\linewidth]{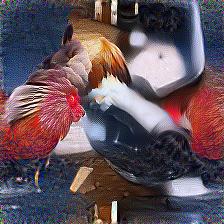}} &
        \fbox{\includegraphics[width=0.95\linewidth]{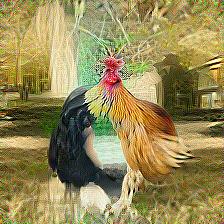}} &
        \fbox{\includegraphics[width=0.95\linewidth]{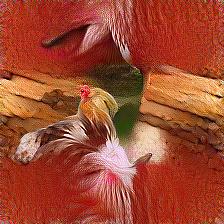}} &
        \fbox{\includegraphics[width=0.95\linewidth]{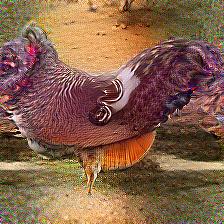}} &
        \fbox{\includegraphics[width=0.95\linewidth]{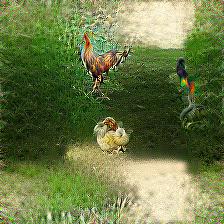}} &
        \fbox{\includegraphics[width=0.95\linewidth]{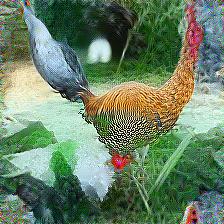}} &
        \fbox{\includegraphics[width=0.95\linewidth]{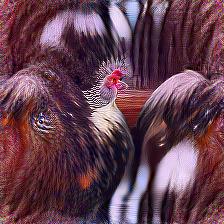}} &
        \fbox{\includegraphics[width=0.95\linewidth]{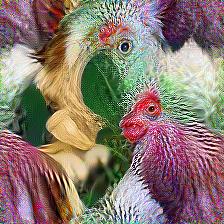}} &
        \fbox{\includegraphics[width=0.95\linewidth]{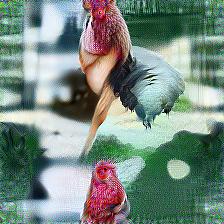}}
        \\
        \multicolumn{10}{c}{\footnotesize $\lambda=100$} \\
        \fbox{\includegraphics[width=0.95\linewidth]{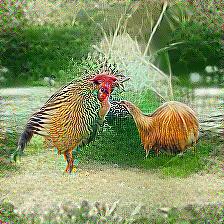}} &
        \fbox{\includegraphics[width=0.95\linewidth]{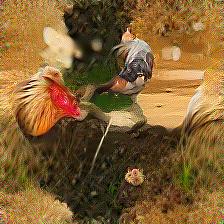}} &
        \fbox{\includegraphics[width=0.95\linewidth]{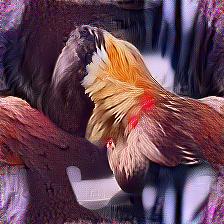}} &
        \fbox{\includegraphics[width=0.95\linewidth]{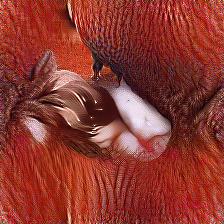}} &
        \fbox{\includegraphics[width=0.95\linewidth]{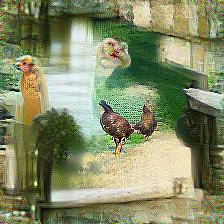}} &
        \fbox{\includegraphics[width=0.95\linewidth]{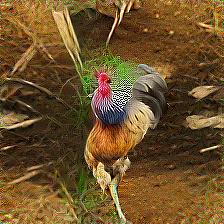}} &
        \fbox{\includegraphics[width=0.95\linewidth]{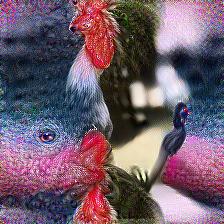}} &
        \fbox{\includegraphics[width=0.95\linewidth]{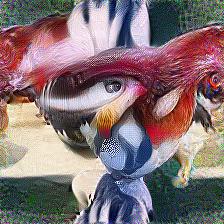}} &
        \fbox{\includegraphics[width=0.95\linewidth]{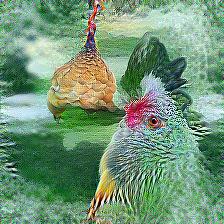}} &
        \fbox{\includegraphics[width=0.95\linewidth]{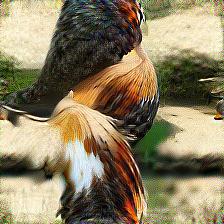}}
        \\
        \multicolumn{10}{c}{\footnotesize $\lambda=1000$} \\

    \end{tabular}
    \end{adjustbox}
    \caption{Visualization of synthetic images in ImageNet-1K with different regularization coefficient $\lambda$}
    \label{fig:Synthetic_Image_ImageNet_compare1}
\end{figure*}

\begin{figure*}[h]
    \centering
    \begin{adjustbox}{width=\linewidth,center}
    \renewcommand{\arraystretch}{1}
    \setlength{\fboxsep}{0pt}
    \begin{tabular}{@{}*{10}{m{0.1\textwidth}@{}}}

        \fbox{\includegraphics[width=0.95\linewidth]{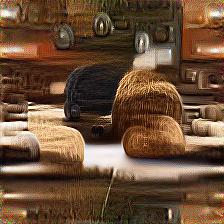}} &
        \fbox{\includegraphics[width=0.95\linewidth]{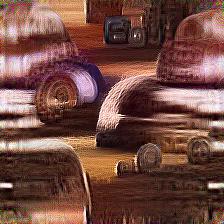}} &
        \fbox{\includegraphics[width=0.95\linewidth]{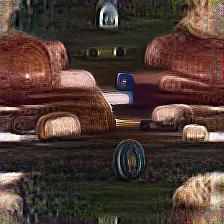}} &
        \fbox{\includegraphics[width=0.95\linewidth]{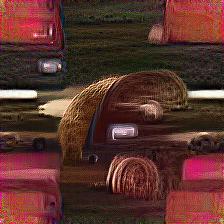}} &
        \fbox{\includegraphics[width=0.95\linewidth]{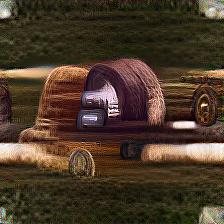}} &
        \fbox{\includegraphics[width=0.95\linewidth]{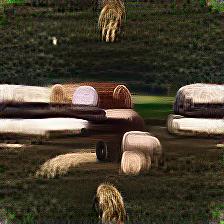}} &
        \fbox{\includegraphics[width=0.95\linewidth]{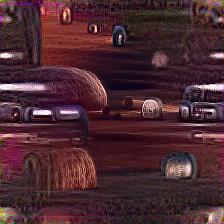}} &
        \fbox{\includegraphics[width=0.95\linewidth]{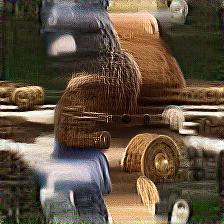}} &
        \fbox{\includegraphics[width=0.95\linewidth]{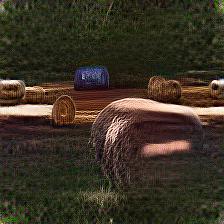}} &
        \fbox{\includegraphics[width=0.95\linewidth]{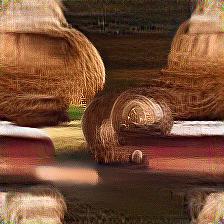}}
        \\
        \multicolumn{10}{c}{\footnotesize SRe$^2$L synthetic images of class Hay (classId: 958)} \\

        \fbox{\includegraphics[width=0.95\linewidth]{figs/Imagenet_our_v2_10ipc/class958_id000.jpg}} &
        \fbox{\includegraphics[width=0.95\linewidth]{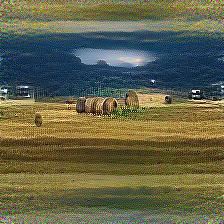}} &
        \fbox{\includegraphics[width=0.95\linewidth]{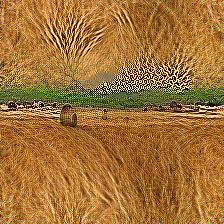}} &
        \fbox{\includegraphics[width=0.95\linewidth]{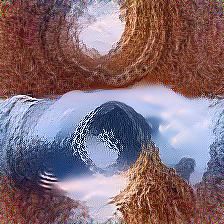}} &
        \fbox{\includegraphics[width=0.95\linewidth]{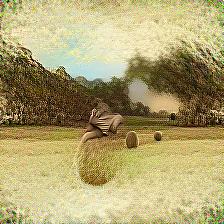}} &
        \fbox{\includegraphics[width=0.95\linewidth]{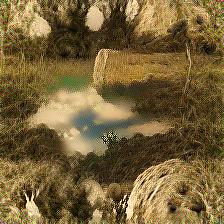}} &
        \fbox{\includegraphics[width=0.95\linewidth]{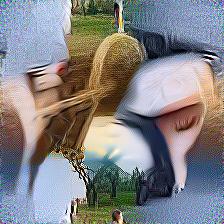}} &
        \fbox{\includegraphics[width=0.95\linewidth]{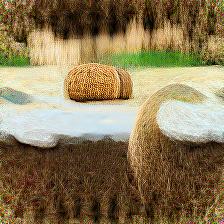}} &
        \fbox{\includegraphics[width=0.95\linewidth]{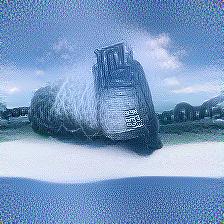}} &
        \fbox{\includegraphics[width=0.95\linewidth]{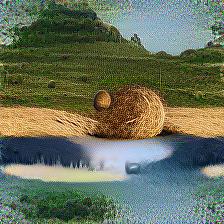}}
        \\
        \multicolumn{10}{c}{\footnotesize Our synthetic images of the class Hay (classId: 958)} \\

        \end{tabular}
    \end{adjustbox}
    \caption{Visualizations of our synthetic images vs. SRe$^2$L baseline synthetic images from ImageNet-1K Hay class (classId: 958).}
    \label{fig:Synthetic_Image_ImageNet_flamingo}

\end{figure*}

\begin{figure*}
    \centering
    \begin{adjustbox}{width=\linewidth,center}
    \renewcommand{\arraystretch}{1}
    \setlength{\fboxsep}{0pt}
    \begin{tabular}{@{}*{10}{m{0.1\textwidth}@{}}}
        \fbox{\includegraphics[width=0.95\linewidth]{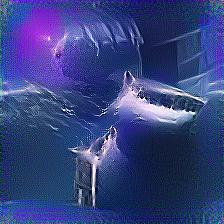}} &
        \fbox{\includegraphics[width=0.95\linewidth]{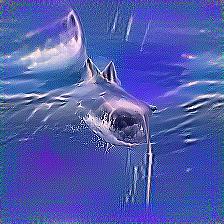}} &
        \fbox{\includegraphics[width=0.95\linewidth]{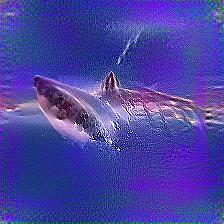}} &
        \fbox{\includegraphics[width=0.95\linewidth]{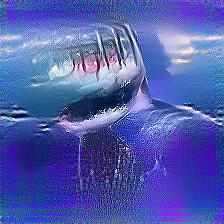}} &
        \fbox{\includegraphics[width=0.95\linewidth]{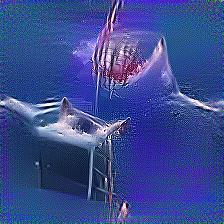}} &
        \fbox{\includegraphics[width=0.95\linewidth]{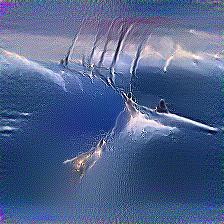}} &
        \fbox{\includegraphics[width=0.95\linewidth]{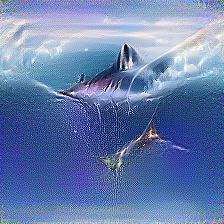}} &
        \fbox{\includegraphics[width=0.95\linewidth]{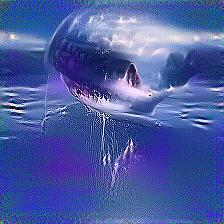}} &
        \fbox{\includegraphics[width=0.95\linewidth]{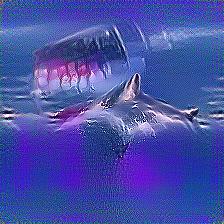}} &
        \fbox{\includegraphics[width=0.95\linewidth]{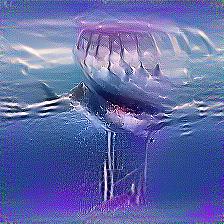}}
        \\
        \multicolumn{10}{c}{\footnotesize SRe$^2$L synthetic images of class White Shark (classId: 002)} \\

        \fbox{\includegraphics[width=0.95\linewidth]{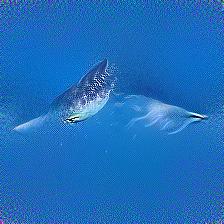}} &
        \fbox{\includegraphics[width=0.95\linewidth]{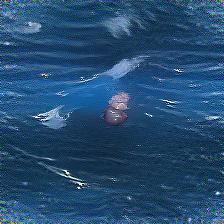}} &
        \fbox{\includegraphics[width=0.95\linewidth]{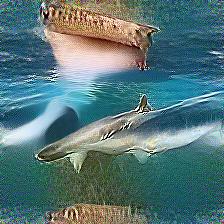}} &
        \fbox{\includegraphics[width=0.95\linewidth]{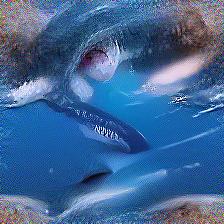}} &
        \fbox{\includegraphics[width=0.95\linewidth]{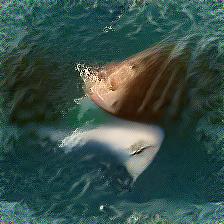}} &
        \fbox{\includegraphics[width=0.95\linewidth]{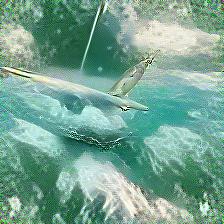}} &
        \fbox{\includegraphics[width=0.95\linewidth]{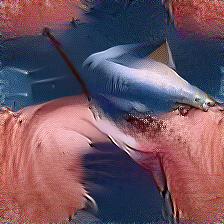}} &
        \fbox{\includegraphics[width=0.95\linewidth]{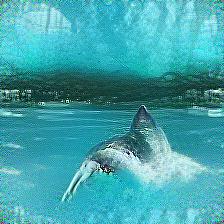}} &
        \fbox{\includegraphics[width=0.95\linewidth]{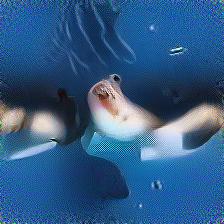}} &
        \fbox{\includegraphics[width=0.95\linewidth]{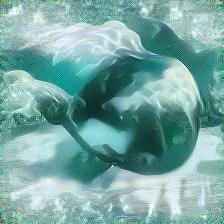}}
        \\
        \multicolumn{10}{c}{\footnotesize Our synthetic images of the class White Shark (classId: 002)} \\[14pt]
        \fbox{\includegraphics[width=0.95\linewidth]{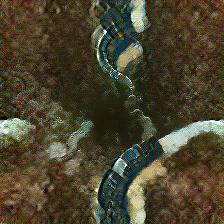}} &
        \fbox{\includegraphics[width=0.95\linewidth]{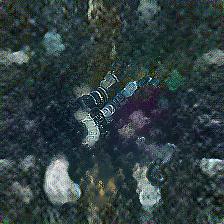}} &
        \fbox{\includegraphics[width=0.95\linewidth]{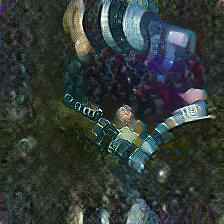}} &
        \fbox{\includegraphics[width=0.95\linewidth]{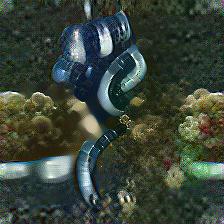}} &
        \fbox{\includegraphics[width=0.95\linewidth]{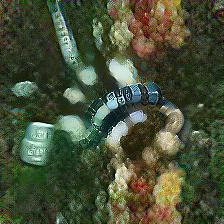}} &
        \fbox{\includegraphics[width=0.95\linewidth]{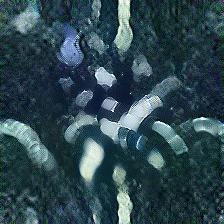}} &
        \fbox{\includegraphics[width=0.95\linewidth]{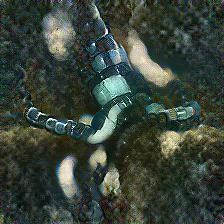}} &
        \fbox{\includegraphics[width=0.95\linewidth]{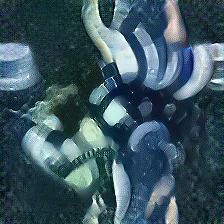}} &
        \fbox{\includegraphics[width=0.95\linewidth]{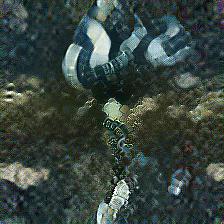}} &
        \fbox{\includegraphics[width=0.95\linewidth]{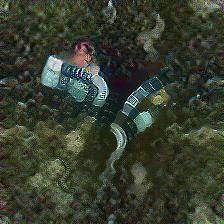}}
        \\
        \multicolumn{10}{c}{\footnotesize SRe$^2$L synthetic images of class Sea Snake (classId: 065)} \\

        \fbox{\includegraphics[width=0.95\linewidth]{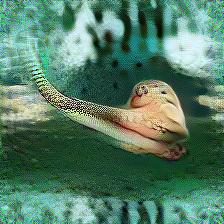}} &
        \fbox{\includegraphics[width=0.95\linewidth]{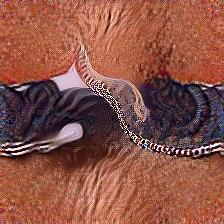}} &
        \fbox{\includegraphics[width=0.95\linewidth]{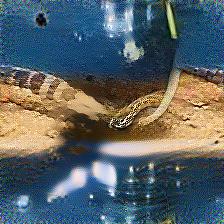}} &
        \fbox{\includegraphics[width=0.95\linewidth]{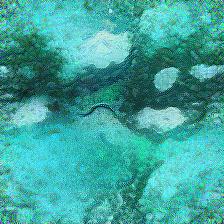}} &
        \fbox{\includegraphics[width=0.95\linewidth]{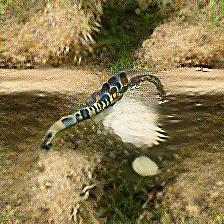}} &
        \fbox{\includegraphics[width=0.95\linewidth]{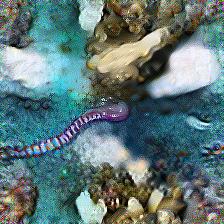}} &
        \fbox{\includegraphics[width=0.95\linewidth]{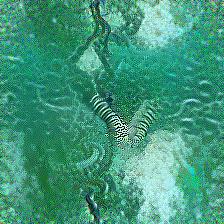}} &
        \fbox{\includegraphics[width=0.95\linewidth]{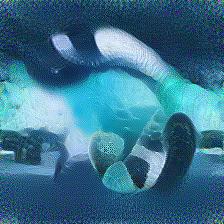}} &
        \fbox{\includegraphics[width=0.95\linewidth]{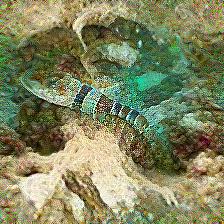}} &
        \fbox{\includegraphics[width=0.95\linewidth]{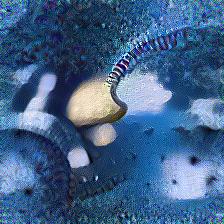}}
        \\
        \multicolumn{10}{c}{\footnotesize Our synthetic images of the class Sea Snake (classId: 065)} \\[14pt]
        \fbox{\includegraphics[width=0.95\linewidth]{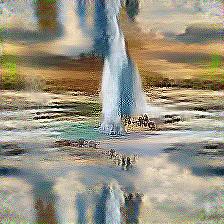}} &
        \fbox{\includegraphics[width=0.95\linewidth]{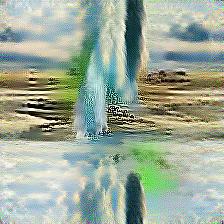}} &
        \fbox{\includegraphics[width=0.95\linewidth]{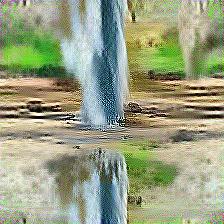}} &
        \fbox{\includegraphics[width=0.95\linewidth]{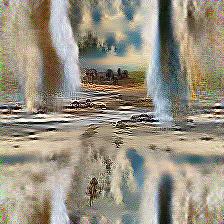}} &
        \fbox{\includegraphics[width=0.95\linewidth]{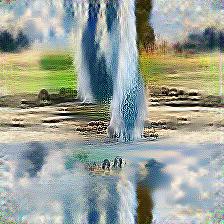}} &
        \fbox{\includegraphics[width=0.95\linewidth]{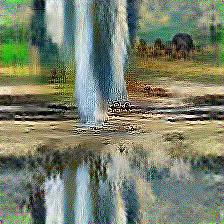}} &
        \fbox{\includegraphics[width=0.95\linewidth]{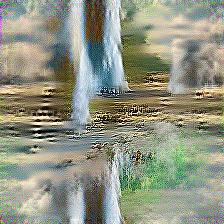}} &
        \fbox{\includegraphics[width=0.95\linewidth]{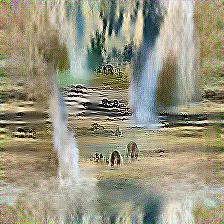}} &
        \fbox{\includegraphics[width=0.95\linewidth]{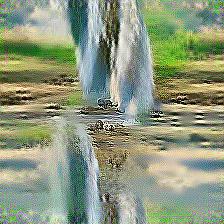}} &
        \fbox{\includegraphics[width=0.95\linewidth]{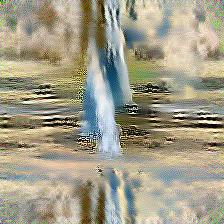}}
        \\
        \multicolumn{10}{c}{\footnotesize SRe$^2$L synthetic images of class Geyser (classId: 974)} \\

        \fbox{\includegraphics[width=0.95\linewidth]{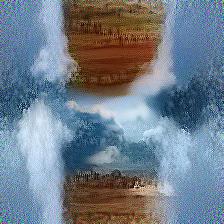}} &
        \fbox{\includegraphics[width=0.95\linewidth]{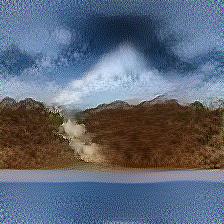}} &
        \fbox{\includegraphics[width=0.95\linewidth]{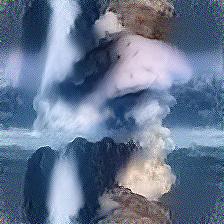}} &
        \fbox{\includegraphics[width=0.95\linewidth]{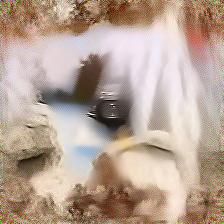}} &
        \fbox{\includegraphics[width=0.95\linewidth]{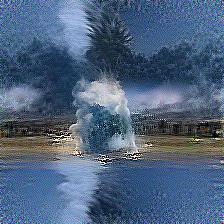}} &
        \fbox{\includegraphics[width=0.95\linewidth]{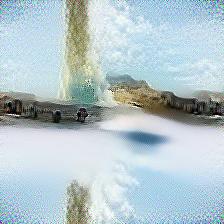}} &
        \fbox{\includegraphics[width=0.95\linewidth]{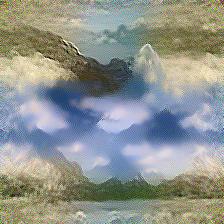}} &
        \fbox{\includegraphics[width=0.95\linewidth]{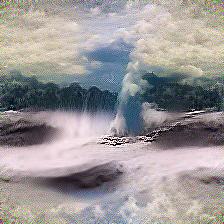}} &
        \fbox{\includegraphics[width=0.95\linewidth]{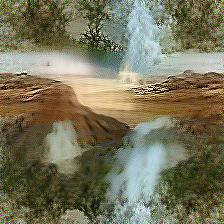}} &
        \fbox{\includegraphics[width=0.95\linewidth]{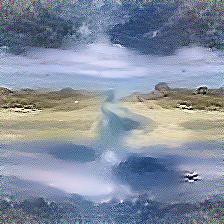}}
        \\
        \multicolumn{10}{c}{\footnotesize Our synthetic images of the class Geyser (classId: 974)} \\[14pt]
        
        \fbox{\includegraphics[width=0.95\linewidth]{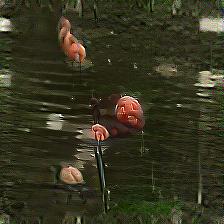}} &
        \fbox{\includegraphics[width=0.95\linewidth]{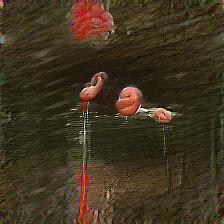}} &
        \fbox{\includegraphics[width=0.95\linewidth]{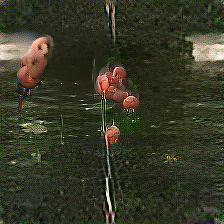}} &
        \fbox{\includegraphics[width=0.95\linewidth]{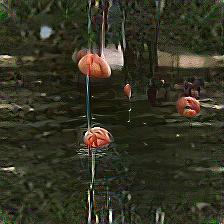}} &
        \fbox{\includegraphics[width=0.95\linewidth]{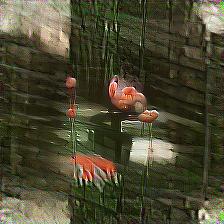}} &
        \fbox{\includegraphics[width=0.95\linewidth]{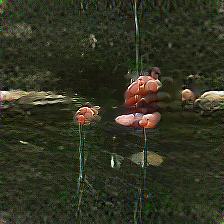}} &
        \fbox{\includegraphics[width=0.95\linewidth]{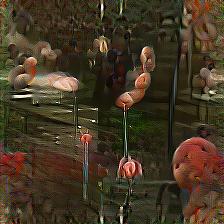}} &
        \fbox{\includegraphics[width=0.95\linewidth]{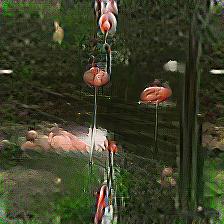}} &
        \fbox{\includegraphics[width=0.95\linewidth]{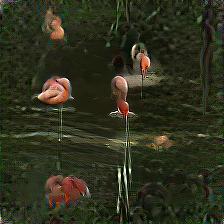}} &
        \fbox{\includegraphics[width=0.95\linewidth]{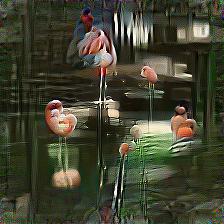}}
        \\
        \multicolumn{10}{c}{\footnotesize SRe$^2$L synthetic images of class Flamingo (classId: 130)} \\

        \fbox{\includegraphics[width=0.95\linewidth]{figs/Imagenet_our_v2_10ipc/class130_id000.jpg}} &
        \fbox{\includegraphics[width=0.95\linewidth]{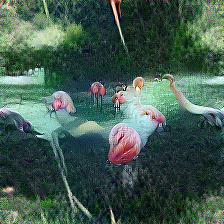}} &
        \fbox{\includegraphics[width=0.95\linewidth]{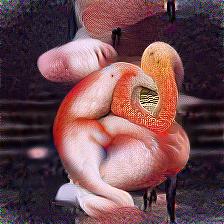}} &
        \fbox{\includegraphics[width=0.95\linewidth]{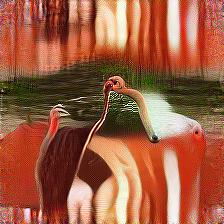}} &
        \fbox{\includegraphics[width=0.95\linewidth]{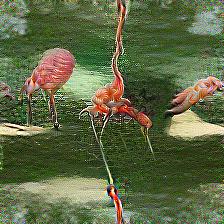}} &
        \fbox{\includegraphics[width=0.95\linewidth]{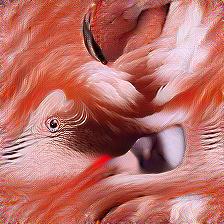}} &
        \fbox{\includegraphics[width=0.95\linewidth]{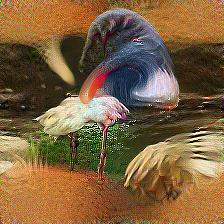}} &
        \fbox{\includegraphics[width=0.95\linewidth]{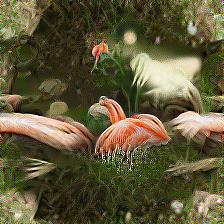}} &
        \fbox{\includegraphics[width=0.95\linewidth]{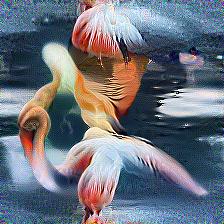}} &
        \fbox{\includegraphics[width=0.95\linewidth]{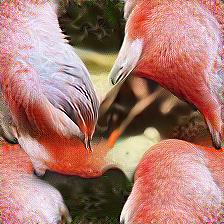}}
        \\
        \multicolumn{10}{c}{\footnotesize Our synthetic images of the class Flamingo (classId: 130)} \\
    \end{tabular}
    \end{adjustbox}
    \caption{Comparison of synthetic images obtained from our method vs. SRe$^2$L on ImageNet-1K in 10 IPC setting. Our method yields synthetic images that better cover the diversity of real images within each class.}
    \label{fig:Synthetic_Image_ImageNet_more}
\end{figure*}

\section{Increased Variety in Synthetic Images}\label{sec: variety}
Visualization of the synthetic images at the pixel level corroborates our finding in Section \ref{sec: tsne} of the main paper, with the ImageNet-1K Hay class being one such example, as shown in Figure \ref{fig:Synthetic_Image_ImageNet_flamingo}. Compared to the SRe$^2$L baseline synthetic images, our method leads to improved variety in both the background and foreground information contained in synthetic images. 
By covering the variety of images in the real data distribution, our method prevents the model from relying on a specific background color or object layout as heuristics for prediction, thus alleviating the potential overfitting problem and improving the generalization of the model. We provide more visualization on three datasets in Appendix \ref{sec: visualizations}.

\section{Visualizations}
\label{sec: visualizations}

We provide visualization of synthetic images from our condensed dataset in the supplementary material. In Figure \ref{fig:visualization-1K}, our observations reveal that the synthetic images produced through our methodology exhibit a remarkable level of semantic clarity, successfully capturing the essential attributes and outlines of the intended class. This illustrates that underscores the fact that our approach yields images of superior quality, which incorporate an abundance of semantic details to enhance validation accuracy and exhibit exceptional visual performance. 

Additionally, Figure \ref{fig:Synthetic_Image_smaller} show our synthetic images on smaller datasets. Figure \ref{fig:Synthetic_Image_ImageNet_compare1} shows the effect of the regularization strength. In Figure \ref{fig:Synthetic_Image_ImageNet_more}, we compare the synthetic data from our method and \citep{yin2023squeeze}. It can be seen that our method enables the synthetic images to convey more diverse foreground and background information, which potentially reduces overfitting and improves the generalization of models trained on those images.

\clearpage

\end{document}